\RequirePackage{fix-cm}
\documentclass{article}

\usepackage{iclr2027_conference,times}

\usepackage{amsmath,amsfonts,bm}

\renewcommand{\eqref}[1]{\textup{(\ref{#1})}}

\def\1{\bm{1}}

\DeclareMathAlphabet{\mathsfit}{\encodingdefault}{\sfdefault}{m}{sl}
\SetMathAlphabet{\mathsfit}{bold}{\encodingdefault}{\sfdefault}{bx}{n}

\usepackage[utf8]{inputenc}
\usepackage[T1]{fontenc}
\usepackage{microtype}
\usepackage{graphicx}
\usepackage{booktabs}
\usepackage{xcolor}
\usepackage{tabularx}
\usepackage{longtable}
\usepackage{array}
\usepackage{amsmath}
\usepackage{amssymb}
\usepackage{mathtools}
\usepackage{amsthm}
\usepackage{enumitem}
\usepackage{subcaption}
\usepackage{float}
\usepackage{placeins}
\usepackage{needspace}
\usepackage{xurl}
\usepackage[bookmarks=false]{hyperref}
\definecolor{referenceblue}{HTML}{2F5B8C}
\hypersetup{
  colorlinks=true,
  linkcolor=referenceblue,
  citecolor=referenceblue,
  urlcolor=referenceblue,
  pdftitle={SGNO: Spectral Generator Neural Operator for Long-Horizon PDE Prediction},
  pdfauthor={Jiayi Li, Penghao Jiang, Baiyu Chen, Yifan Duan, Hira Saleem, Zhaonan Wang, Piotr Koniusz, Flora D. Salim}
}
\usepackage[capitalize,noabbrev]{cleveref}

\graphicspath{{figures/}}


\theoremstyle{plain}
\newtheorem{theorem}{Theorem}[section]
\newtheorem{proposition}[theorem]{Proposition}

\theoremstyle{definition}

\theoremstyle{remark}

\title{SGNO: Spectral Generator Neural Operator for Long-Horizon PDE Prediction}

\author{\begin{minipage}{\dimexpr\textwidth-2\tabcolsep\relax}\centering\normalfont
Jiayi Li\textsuperscript{1}, Penghao Jiang\textsuperscript{1}, Baiyu Chen\textsuperscript{1}, Yifan Duan\textsuperscript{1},\\
Hira Saleem\textsuperscript{1}, Zhaonan Wang\textsuperscript{2}, Piotr Koniusz\textsuperscript{1,3,$\dagger$}, Flora D. Salim\textsuperscript{1,$\dagger$}\\[0.6em]
{\small \textsuperscript{1}University of New South Wales\\
\textsuperscript{2}New York University\\
\textsuperscript{3}Commonwealth Scientific and Industrial Research Organisation}
\end{minipage}}

\iclrfinalcopy

\newcommand{\ms}[2]{\ensuremath{#1}\,{\fontsize{8}{9}\selectfont\ensuremath{\pm #2}}}
\begin{document}

\maketitle
\lhead{}
\renewcommand{\headrulewidth}{0pt}
\begingroup
\renewcommand{\thefootnote}{\fnsymbol{footnote}}
\interfootnotelinepenalty=10000
\footnotetext[2]{\raggedright Corresponding authors: Piotr Koniusz (\href{mailto:piotr.koniusz@unsw.edu.au}{piotr.koniusz@unsw.edu.au}) and Flora D. Salim (\href{mailto:flora.salim@unsw.edu.au}{flora.salim@unsw.edu.au}).}
\endgroup

\begin{abstract}
Neural operators efficiently simulate dynamical systems by repeatedly
predicting the next state. However, small errors at each step can accumulate,
causing long-term predictions to drift from the true trajectory.
Long-term error reflects both the propagation of earlier prediction errors
and new errors introduced at each step. How to reduce these errors is
therefore a critical question.
We propose the Spectral Generator Neural Operator (SGNO),
which combines a linear estimate of the next state with a learned correction.
The linear component uses consecutive states in the training data to fit
a map from the current state to the next. A neural network corrects
this prediction by learning changes that the linear model does not capture,
including nonlinear interactions. Both components are trained jointly
toward the same next-state target.
Experiments across 13 periodic PDE configurations in one to three dimensions show that SGNO accurately learns state evolution on linear tasks and reduces 100-step rollout error by 13-74\% over the strongest baseline under the main comparison protocol on each nonlinear task. Longer-horizon extrapolation tests show lower late-window prediction errors on several nonlinear tasks. These results show that combining fitted propagation with learned state-dependent correction can improve long-horizon autoregressive prediction on the evaluated PDE tasks.
Code is available at
\url{https://github.com/cruiseresearchgroup/SGNO}.

\end{abstract}

\section{Introduction}
\label{sec:introduction}

Neural operators learn maps between physical fields, providing efficient
surrogates for dynamical systems governed by partial differential equations
(PDEs)~\citep{lu2021deeponet,kovachki2023neuraloperator}.
However, when predictions are repeatedly used as inputs to subsequent steps,
small errors can accumulate and cause the predicted trajectory to drift
from the true evolution. Improving long-horizon prediction accuracy remains
an important research problem. Toward this goal,
researchers have improved network architectures and training
procedures~\citep{li2021fno,zhang2024sinenet,brandstetter2022message},
refined predictions iteratively~\citep{lippe2023pderefiner}, and constrained
sensitivity to input perturbations~\citep{mccabe2023towards,park2024jacobian}.

\begin{figure}[t]
\centering
\includegraphics[width=\linewidth]{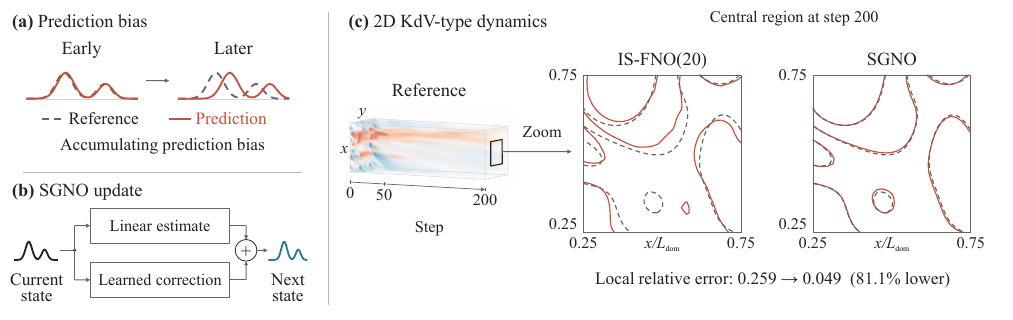}
\caption{\textbf{SGNO overview and long-horizon prediction example.}
(a) Schematic accumulation of prediction bias.
(b) SGNO combines a linear estimate with a learned correction.
(c) An additional 2D KdV-type example at step 200.
Dashed gray and solid red contours denote reference and prediction.
In the outlined central region, relative $L^2$ error is 0.259 for IS-FNO(20)
and 0.049 for SGNO. Coordinates are normalized by domain side length
$L_{\mathrm{dom}}$. The 20-state training objective and this example's
protocol are given in Appendix~\ref{app:phase_example}.}
\label{fig:intro-motivation}
\end{figure}

These advances motivate a closer look at how prediction errors accumulate.
Controlling the amplification of existing errors is important, but even
without such amplification, an inaccurate description of the dynamics can
introduce new errors at every step. Consider a translating wave: if a model
moves it slightly too far at each step, the previous position error is
retained and a new position error is added. The predicted location can
therefore drift from the true location even when the wave amplitude remains
accurate~\citep{williams2013amplitude}, as illustrated in
Figure~\ref{fig:intro-motivation}(a).

This example motivates attention to both the propagation of existing errors
and the accuracy of the state change predicted at each step. Our starting
point is to use the evolution information already present in adjacent
training states to construct a structured initial estimate of the next
state, and then learn the changes that this estimate leaves undescribed.
This leads to the linear-estimate and learned-correction structure shown in
Figure~\ref{fig:intro-motivation}(b).

We propose the \emph{Spectral Generator Neural Operator} (SGNO) based on
this construction. The linear component expresses the physical field as
wave components of different frequencies using a Fourier representation.
It estimates how each component changes from adjacent training states,
while constraining its amplification~\citep{baddoo2023pidmd}.
The correction network learns changes that the linear estimate does not
capture, including nonlinear interactions. Each correction layer combines
existing features with newly computed nonlinear features, using a shared
set of parameters to link the weights of these two contributions.
The correction output is initialized to zero, so the predictor starts from
the fitted map. Both components are then trained jointly toward the
same next-state target~\citep{rodrigues2021resdmd}.
Experiments show that SGNO can reduce long-horizon prediction errors,
bringing predicted trajectories closer to the reference evolution.
Figure~\ref{fig:intro-motivation}(c) gives an illustrative comparison.

Our main contributions are:
\begin{enumerate}[leftmargin=*,nosep]

\item \textbf{Error accumulation beyond amplification.}
We analyze how existing prediction errors propagate and how errors in the learned state update can accumulate under recursive prediction.
A phase-evolution example shows that even non-amplifying updates can produce
growing trajectory error, motivating attention to state-update accuracy
alongside error amplification.

\item \textbf{Fitted propagation with learned correction.}
We construct SGNO by fitting a gain-constrained Fourier map to adjacent
training states and pairing it with a neural correction for state-dependent
changes, including nonlinear interactions.
The correction starts from zero, and both components are trained jointly
toward the same next-state target.

\item \textbf{Controlled evaluation across dynamics and horizons.}
We evaluate SGNO on 13 periodic PDE configurations with five training
runs and recursive prediction through 200 steps.
Component ablations, spectral diagnostics, and late-window analyses
characterize the roles of fitted propagation, learned correction, and joint
adaptation under recursive prediction.

\end{enumerate}


\FloatBarrier

\section{Related Work}
\label{sec:related_work}

\paragraph{Neural operators and spectral representations.}
Neural operators learn mappings between functions, providing a common
framework for learning partial differential equation (PDE) solution maps
\citep{lu2021deeponet,kovachki2023neuraloperator}.
DeepONet uses separate networks for input functions and output
coordinates~\citep{lu2021deeponet}; the Fourier neural operator (FNO)
learns Fourier-space interactions~\citep{li2021fno}.
The convolutional neural operator (CNO) uses multiscale convolutions for function-space
learning~\citep{raonic2023convolutional}.

\paragraph{Long-horizon prediction and error control.}
Recursive prediction exposes a model to its own imperfect outputs.
Training with perturbed states or model-generated inputs addresses this
change in the input distribution
\citep{sanchezgonzalez2020learning,brandstetter2022message,ye2025rno}.
Multi-step objectives also evaluate the consequences of repeated updates,
as in the inverse-scattering-inspired Fourier neural operator
(IS-FNO)~\citep{yu2025isfno}.
PDE-Refiner improves predictions through iterative refinement, including
low-amplitude components that affect later evolution
\citep{lippe2023pderefiner}.
Other approaches constrain model sensitivity
\citep{mccabe2023towards} or regularize long-term state distributions
\citep{schiff2024dyslim}.
These methods address different aspects of recursive error.
They motivate considering both how errors propagate and how accurately
each update describes the state change.

\paragraph{Structured dynamics and learned correction.}
Modeling linear evolution and nonlinear changes separately is a common
approach to constructing time-dependent predictors.
Exponential integrators use this division to advance dynamical systems
\citep{cox2002etd,hochbruck2010exponential}, and structured neural
ordinary differential equations (ODEs)
incorporate it into learned dynamics
\citep{linot2023stabilized,loya2025structurepreserving}.
The stable spectral neural operator (SSNO) learns a continuous-time
generator in physical Fourier coordinates
and combines it with nonlinear terms through integrating-factor
Runge--Kutta updates~\citep{zhang2025ssno}.
IS-FNO and latent structured spectral propagators (SSP) construct spectral
evolution in learned representations
and combine it with nonlinear updates~\citep{yu2025isfno,lu2026ssp}.

A related line of work estimates linear evolution directly from observations.
Dynamic mode decomposition (DMD) fits a linear map from state
pairs~\citep{schmid2010dmd}; physics-informed DMD (piDMD) exploits periodic
translation invariance to perform this regression
independently for each Fourier mode~\citep{baddoo2023pidmd}.
ResDMD augments a DMD-fitted model with a neural residual and jointly
trains the propagation and correction components
\citep{rodrigues2021resdmd}.
SGNO fits a constrained affine discrete-time map from adjacent training
states in the fixed Fourier representation of the observed field, and
jointly trains this map with a spectral correction network.
Within each correction layer, a shared parameter set links the transmission
of existing features to the addition of nonlinear features.

\section{Problem Setting}
\label{sec:problem_setting}
\label{subsec:problem_setting}

We learn a next-state predictor for scalar PDE fields on periodic grids.
In $d$ spatial dimensions, let $\mathbf u^{[t]}\in\mathbb R^{n_g}$
collect the grid values, with $n_g=\prod_{j=1}^dN_j$ and $N_j$ points
along axis $j$; $h$ denotes this spatial grid.
The index $t$ counts stored states separated by the interval $\Delta t$.
Training data comprise $M$ adjacent state pairs
$\mathcal D=\{(\mathbf u_j,\mathbf u'_j)\}_{j=1}^{M}$, where
$\mathbf u'_j$ is the next observed reference state, including spatial
sampling when used (Appendix~\ref{app:data_configuration}). The predictor
$f_{\boldsymbol\theta}$, with trainable parameters $\boldsymbol\theta$,
learns this one-step prediction relation without the governing PDE
coefficients. Let $\widetilde{\mathbf u}^{[t]}$ denote the predicted state
at step $t$. Starting from an unseen initial state, the model recursively
uses its own predictions:
\begin{equation}
 \widetilde{\mathbf u}^{[t+1]}=f_{\boldsymbol\theta}(\widetilde{\mathbf u}^{[t]}),
 \qquad \widetilde{\mathbf u}^{[0]}=\mathbf u^{[0]}.
 \label{eq:rollout}
\end{equation}
Training trajectories cover 50 transitions; testing predicts 200 steps
at the same grid and output interval, without reference-state resets.
Steps 51--200 lie beyond the training trajectories' time span.
The 50 transitions supply adjacent pairs for one-step training.
The SGNO 2D initialization was selected using separate 200-step
development trajectories (Appendix~\ref{app:model_configuration}).

The data have a fixed, approximately zero spatial mean
(Appendix~\ref{app:mean_scope}). Vectors are bold lowercase, matrices bold
uppercase, and functions unbold lowercase; scalar coefficients are unbold.
Arrays are vectorized; hats denote Fourier coefficients and
tildes denote predictions.

\section{Method}
\label{sec:method}
\begin{figure}[!t]
\centering
\includegraphics[width=\linewidth]{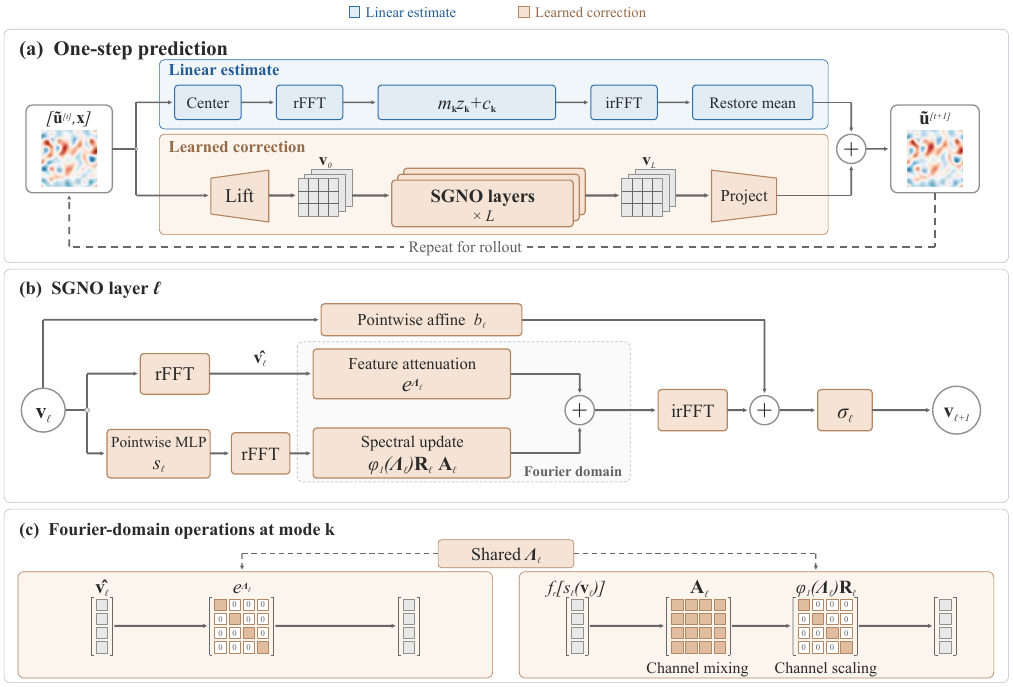}
\caption{\textbf{SGNO architecture.} (a) Linear estimate (blue) plus learned
correction (orange); dashed loop: recursive prediction.
(b) An SGNO layer combines pointwise and spectral updates.
(c) A shared generator links feature attenuation and nonlinear addition.
Here $z_{\mathbf k}=[f_r(\mathbf q)]_{\mathbf k}$ and
$\widehat{\mathbf v}_\ell=f_r(\mathbf v_\ell)$. Insets show a 2D KdV-type
state and prediction on a common color scale.}
\label{fig:sgno_architecture}
\end{figure}

Figure~\ref{fig:sgno_architecture}(a) combines a fitted Fourier estimate
with a learned correction for next-state prediction.

\subsection{Repeated errors and complementary prediction roles}
\label{subsec:learning_evolution}\label{subsec:time_evolution_structure}
\label{subsec:physical_construction}
Let $\mathbf e^{[t]}=\widetilde{\mathbf u}^{[t]}-\mathbf u^{[t]}$ denote
the prediction error at step $t$.
Adding and subtracting the prediction from the reference state gives
\begin{equation}
\mathbf e^{[t+1]}=
\underbrace{f_{\boldsymbol\theta}(\widetilde{\mathbf u}^{[t]})-
f_{\boldsymbol\theta}(\mathbf u^{[t]})}_{\text{effect of input error}}
+\underbrace{f_{\boldsymbol\theta}(\mathbf u^{[t]})-
\mathbf u^{[t+1]}}_{\text{error in the learned update}}.
\label{eq:rollout_error_decomposition}
\end{equation}
Controlling sensitivity to an inaccurate input addresses the first term;
the second introduces new errors at successive steps. A phase error gives
a simple example. For a nonzero Fourier coefficient $z$, let $\omega$ be
the true phase advance per step and $\varepsilon$ its prediction error.
The true and learned multipliers $e^{i\omega}$ and
$e^{i(\omega+\varepsilon)}$, with $i^2=-1$, both preserve magnitude.
After $n$ prediction steps,
\begin{equation}
\frac{|e^{in(\omega+\varepsilon)}z-e^{in\omega}z|}{|z|}
=2\left|\sin\frac{n\varepsilon}{2}\right|
\simeq n|\varepsilon|\quad(n|\varepsilon|\ll1).
\label{eq:phase_accumulation}
\end{equation}
Thus accurate phase advance matters even without amplification
\citep{williams2013amplitude}. Appendix~\ref{app:phase_accumulation_derivation}
derives the identity and approximation.

For periodic linear dynamics that act uniformly across space, complex
Fourier multipliers describe amplitude change and phase advance. Adjacent
observations allow these changes to be estimated directly. Independent
modes cannot express nonlinear frequency interactions, so a neural
correction learns the remaining changes, including errors left by the fit.
Both branches jointly determine the propagation and update errors in
Equation~\eqref{eq:rollout_error_decomposition}.

For input $\mathbf w\in\mathbb R^{n_g}$, let
$\mathbf P_0\in\mathbb R^{n_g\times n_g}$ replace each grid entry by the
spatial mean. With $\mathbf I$ the identity matrix, define the centering
matrix $\mathbf Q=\mathbf I-\mathbf P_0$ and centered state
$\mathbf q=\mathbf Q\mathbf w$. The linear estimate $a_{\boldsymbol\eta}$,
with parameters $\boldsymbol\eta$, and neural correction
$r_{\boldsymbol\theta}$ form
\begin{equation}
f_{\boldsymbol\theta}(\mathbf w)=\mathbf P_0\mathbf w+
\mathbf Q\!\left[a_{\boldsymbol\eta}(\mathbf q)+
r_{\boldsymbol\theta}(\mathbf q)\right].
\label{eq:physical_decomposition}
\end{equation}
Here $\boldsymbol\theta$ includes $\boldsymbol\eta$, and both branches
receive the centered state $\mathbf q$. In the diagram, Restore mean
removes the estimate's mean and restores the input mean;
Project returns a zero-mean correction. Together these operations implement
the fixed-mean setting (Appendix~\ref{app:mean_scope}). Their sum forms
an empirical map from the current state to the next.
During rollout, $\mathbf w=\widetilde{\mathbf u}^{[t]}$ and the sum becomes
the next input.

\subsection{Estimating a Fourier map from adjacent states}
\label{subsec:propagator_fit}
Following Fourier-diagonal regression~\citep{baddoo2023pidmd}, we fit one
complex multiplier per frequency. An input-independent offset permits
a fixed spatial contribution, so the linear estimate is formally affine.
Center each training pair as $\mathbf q_j=\mathbf Q\mathbf u_j$ and
$\mathbf q'_j=\mathbf Q\mathbf u'_j$. For the integer Fourier index
$\mathbf k=(k_1,\ldots,k_d)$, define
$x_j=[\widehat{\mathbf q}_j]_{\mathbf k}$ and
$y_j=[\widehat{\mathbf q}'_j]_{\mathbf k}$.
The multiplier $m_{\mathbf k}$ describes the change of this coefficient;
$c_{\mathbf k}$ is its offset. Multipliers cover the full Fourier grid,
while offsets are learned only within a retained frequency band $\mathcal K$.
The allowed offset set is $\mathcal C_{\mathbf k}=\mathbb C$ inside
$\mathcal K$ and $\{0\}$ outside. Using all $M$ pairs, we solve
\begin{equation}
(m_{\mathbf k}^{0},c_{\mathbf k}^{0})=
\arg\min_{|m|\leq1,\ c\in\mathcal C_{\mathbf k}}
\sum_{j=1}^{M}|y_j-mx_j-c|^2.
\label{eq:affine_fit}
\end{equation}
Superscript $0$ denotes initialization. The gain constraint allows arbitrary
phase change while limiting amplification.

The fit has a closed-form solution: project the least-squares multiplier
onto the unit disk without changing its angle, then recompute the offset.
Outside $\mathcal K$, the offset remains zero.
Appendix~\ref{app:propagation_construction} gives the formula and proof
(Proposition~\ref{prop:affine_fit}); Appendix~\ref{app:parameterization}
specifies low-energy modes.

Collect the coefficients in vectors $\mathbf m_{\boldsymbol\eta}$ and
$\mathbf c_{\boldsymbol\eta}$. Let $f_r$ be the real fast Fourier transform
(rFFT) and $f_r^{-1}$ its inverse, with $\odot$ denoting elementwise multiplication. The projection
$p_H$ enforces the Fourier symmetry required for a real-valued output:
\begin{equation}
a_{\boldsymbol\eta}(\mathbf q)=f_r^{-1}\!\left(
p_H[\mathbf m_{\boldsymbol\eta}\odot f_r(\mathbf q)+
\mathbf c_{\boldsymbol\eta}]\right).
\label{eq:main_spectral_map}
\end{equation}
In the blue branch, Center forms $\mathbf q$ and rFFT obtains each
coefficient $z_{\mathbf k}=[f_r(\mathbf q)]_{\mathbf k}$. The update is
$m_{\mathbf k}z_{\mathbf k}+c_{\mathbf k}$. The inverse transform and
Restore mean return the next-state estimate. Fitting uses only observed
training pairs and does not require PDE coefficients. Joint training uses
$m_{\mathbf k}=\exp(-\rho_{\mathbf k}^{2}+i\omega_{\mathbf k})$, where the
real parameters $\rho_{\mathbf k}$ and $\omega_{\mathbf k}$ control
magnitude and phase advance, respectively. This preserves the modal gain bound.

\subsection{Learning the correction}
\label{subsec:bounded_diagonal_propagation}\label{subsec:diagonal_multiplier_update}
\label{subsec:one_step_sgno}\label{subsec:spectral_update_and_local}
The orange correction branch consists of Lift, $L$ SGNO layers, and Project.
In the diagram, Lift groups input centering with the mapping of
$[\mathbf q,\mathbf x]$ to a feature field $\mathbf v_0$, where
$\mathbf x$ denotes grid coordinates and each point has $C$ feature channels.
Layer $\ell=0,\ldots,L-1$ maps $\mathbf v_\ell$ to $\mathbf v_{\ell+1}$;
parameters are separate across layers.

Figure~\ref{fig:sgno_architecture}(b) expands one layer. It combines a
pointwise affine channel map $b_\ell$ with a Fourier-space update. The
Fourier update weights the input features and adds new features computed
by a pointwise multilayer perceptron (MLP) $s_\ell$.
At each retained frequency, a real diagonal matrix
$\boldsymbol\Lambda_\ell$ with nonpositive entries determines the weights
of these two contributions. We call it the \emph{spectral generator}:
$e^{\boldsymbol\Lambda_\ell}$ weights the input features, while
$\phi_1(\boldsymbol\Lambda_\ell)$ weights the nonlinear contribution, with
$\phi_1(z)=(e^z-1)/z$ and $\phi_1(0)=1$.
These functions act on the diagonal entries of $\boldsymbol\Lambda_\ell$.
The complex matrix $\mathbf A_\ell$ mixes channels and the real diagonal
matrix $\mathbf R_\ell$ scales them; all three matrices are $C\times C$.
The mask $\boldsymbol\mu$ is one at retained modes and zero elsewhere.
The activation $\sigma_\ell$ is the Gaussian error linear unit (GELU)
except in the final layer, where it is the identity. Let
$\widehat{\mathbf h}_\ell$ denote the intermediate spectrum. Suppressing
frequency arguments, the layer computes
\begin{equation}
\begin{aligned}
\widehat{\mathbf h}_\ell
&=e^{\boldsymbol\Lambda_\ell}f_r(\mathbf v_\ell)
+\phi_1(\boldsymbol\Lambda_\ell)\mathbf R_\ell\mathbf A_\ell
f_r(s_\ell(\mathbf v_\ell)),\\
\mathbf v_{\ell+1}
&=\sigma_\ell\!\left(b_\ell(\mathbf v_\ell)+
f_r^{-1}p_H[\boldsymbol\mu\odot\widehat{\mathbf h}_\ell]\right).
\end{aligned}
\label{eq:main_corrector_layer}
\end{equation}
Figure~\ref{fig:sgno_architecture}(c) expands the two spectral terms.
Both use the same $\boldsymbol\Lambda_\ell$, and the diagonal factors
$e^{\boldsymbol\Lambda_\ell}$ and $\phi_1(\boldsymbol\Lambda_\ell)$ have
entries in $(0,1]$. Appendix~\ref{app:feature_update_derivation} derives
these factors and their zero-attenuation limit by integrating an auxiliary
linear feature equation with a fixed source. This step processes features;
the outer $m_{\mathbf k}$ updates observed states. Each prediction evaluates
all $L$ layers; pointwise nonlinear maps enable cross-frequency interactions.

Project returns the filtered, zero-mean correction
$\mathbf Qr_{\boldsymbol\theta}(\mathbf q)$; Appendix~\ref{app:parameterization}
specifies its readout and zero-input subtraction. Zero-initializing the
readout makes the initial predictor equal the fitted map. Both branches
then minimize next-state mean squared error (MSE), using grid-normalized
norm $\|\cdot\|_h$ and Euclidean norm $\|\cdot\|_2$:
\begin{equation}
\min_{\boldsymbol\theta}\ \frac1M\sum_{j=1}^{M}
\|f_{\boldsymbol\theta}(\mathbf u_j)-\mathbf u'_j\|_h^2,
\qquad \|\mathbf z\|_h^2=n_g^{-1}\|\mathbf z\|_2^2.
\label{eq:joint_training}
\end{equation}
\FloatBarrier

\section{Experiments}
\label{sec:experiments}\label{subsec:experimental_protocol}
\label{subsec:standard_baselines}\label{subsec:task_structure}
We test recovery of linear updates, the contribution of learned
interactions on nonlinear tasks, and accumulated error during extended
prediction. Ablations and spectral diagnostics examine these roles.

\paragraph{Tasks and training.}
We use thirteen periodic scalar-field APEBench configurations
\citep{koehler2024apebench}: nine linear tasks covering transport,
diffusion, dispersion, and their combinations, and four nonlinear tasks:
Korteweg--de Vries-type (KdV), Kuramoto--Sivashinsky (KS), Burgers,
and Kolmogorov vorticity.
Grids are $160$, $64^2$, and $32^3$ in 1D/2D/3D. Each task provides
50 trajectories of 50 transitions, or 2,500 adjacent training pairs.
Main-table models use 10,000 updates and five training seeds.
Testing predicts 200 steps from 30 unseen initial states in 1D/2D and
ten in 3D, without reference resets. Parameters, grids, output intervals,
and the approximately zero mean are fixed, matching
Section~\ref{sec:problem_setting}. Targets follow the specified reference
protocol; Appendix~\ref{app:data_configuration} gives task settings and
reference time-discretization checks.

\paragraph{Baselines and metrics.}
FNO provides the standard spectral neural-operator baseline
\citep{li2021fno}, CNO tests convolutional representations
\citep{raonic2023convolutional}, the Koopman neural operator (KNO)
tests Koopman dynamics \citep{xiong2024kno}, and IS-FNO tests
spectral evolution \citep{yu2025isfno}.
All use one-step state MSE except KNO's prediction--reconstruction loss.
IS-FNO(1) is the one-step adaptation; its 20-state objective is evaluated
in Appendix~\ref{app:additional_protocols}.
For each seed, GMean100 is the geometric mean over steps 1--100 of
trajectory-averaged normalized root mean squared error (nRMSE).
Main tables report five-seed medians; Appendices~\ref{app:full_comparison}
and \ref{app:ablation_results} give each seed and mean $\pm$ sample
standard deviation (SD).
SGNO and IS-FNO use full 32-bit precision (FP32); the other baselines retain
recorded evaluations (Appendix~\ref{app:metrics}).

\subsection{Main results}
\label{subsec:main_results}
\begin{table}[!b]
\centering
\caption{\textbf{Main prediction accuracy.} Median GMean100 over five seeds; lower is better. Bold: lowest median. All seed values and mean $\pm$ SD: Appendix~\ref{app:full_comparison}.}
\label{tab:main-comparison}
\begingroup\small
\setlength{\tabcolsep}{2.3pt}\renewcommand{\arraystretch}{1.14}
\begin{tabular*}{\linewidth}{@{\extracolsep{\fill}}lrrrrr@{}}
\toprule
Task & FNO & CNO & KNO & IS-FNO(1) & SGNO \\
\midrule
\multicolumn{6}{l}{\textit{Linear dynamics}} \\
Advection (1D) & $0.0236$ & $0.854$ & $0.0280$ & $\mathrm{NF}_{1/5}$ & $\mathbf{2.54\mathrm{e}{-6}}$ \\
Diffusion (1D) & $0.0198$ & $0.207$ & $0.0274$ & $1.64\mathrm{e}{-4}$ & $\mathbf{8.12\mathrm{e}{-6}}$ \\
Dispersion (1D) & $0.0322$ & $0.265$ & $0.0277$ & $4.87\mathrm{e}{-5}$ & $\mathbf{5.01\mathrm{e}{-6}}$ \\
Mixed dispersion (2D) & $0.186$ & $0.0516$ & $0.167$ & $0.00122$ & $\mathbf{6.06\mathrm{e}{-6}}$ \\
Axis diffusion (2D) & $0.0804$ & $0.309$ & $0.0172$ & $0.00354$ & $\mathbf{7.74\mathrm{e}{-6}}$ \\
Rotated diffusion (2D) & $0.0783$ & $0.319$ & $0.0461$ & $0.0123$ & $\mathbf{7.57\mathrm{e}{-6}}$ \\
Diagonal diffusion (3D) & $0.166$ & $0.331$ & $0.0198$ & --- & $\mathbf{5.80\mathrm{e}{-6}}$ \\
Unbalanced advection (3D) & $0.784$ & $0.134$ & $0.0432$ & --- & $\mathbf{5.77\mathrm{e}{-6}}$ \\
Advection--diffusion (3D) & $0.144$ & $1.44$ & $0.0171$ & --- & $\mathbf{7.11\mathrm{e}{-6}}$ \\
\midrule
\multicolumn{6}{l}{\textit{Nonlinear dynamics}} \\
KdV (1D) & $0.194$ & $0.473$ & $0.0908$ & $0.0466$ & $\mathbf{0.0122}$ \\
KS (1D) & $0.516$ & $0.360$ & $0.307$ & $0.0800$ & $\mathbf{0.0512}$ \\
Burgers (1D) & $0.0185$ & $0.293$ & $0.0196$ & $0.00584$ & $\mathbf{0.00334}$ \\
Kolmogorov (2D) & $0.666$ & $0.494$ & $0.976$ & $0.552$ & $\mathbf{0.430}$ \\
\bottomrule
\end{tabular*}\endgroup
\par\smallskip\begin{minipage}{\linewidth}\footnotesize
NF$_{1/5}$: one non-finite seed in steps 1--100; ---: not evaluated.
\end{minipage}
\end{table}

Table~\ref{tab:main-comparison} uses two blocks because the tasks test
different aspects of the construction. On linear tasks, SGNO reaches
median GMean100 of $2.54\times10^{-6}$--$8.12\times10^{-6}$.
For these periodic constant-coefficient systems, the reference update
acts independently on Fourier modes and belongs to the fitted map's
representational family. Adjacent state pairs reveal how each mode
changes, so regression can identify multipliers on modes excited by the
training data. No analytic solution or PDE coefficient is supplied to
the predictor. The small errors therefore reflect recovery of the
observed numerical update, including its amplitude changes and phase
advances. Before neural training, fitted-only GMean100 is already
$6.20\times10^{-6}$--$1.60\times10^{-5}$
(Appendix~\ref{app:linear_recovery}, Table~\ref{tab:linear-fitted-only}).

The nonlinear block tests the complete estimate--correction predictor.
Here a single multiplier per mode cannot describe all changes because
the evolution depends on interactions between modes. SGNO has the lowest
median among the main baselines on all four tasks, reducing GMean100 by
73.8\%, 36.0\%, 42.8\%, and 13.0\% relative to each task's strongest
main baseline on KdV, KS, Burgers, and Kolmogorov.
For example, on KS, median GMean100 is 0.516 for FNO,
0.0800 for IS-FNO(1), and 0.0512 for SGNO.
SGNO also has lower mean GMean100 than IS-FNO(20) on the three 1D
nonlinear tasks. At higher capacity and training budget, Refiner-L achieves
lower error on KS (Appendix~\ref{app:additional_protocols}).

\subsection{Ablation study}
\label{subsec:attribution}
Table~\ref{tab:main-ablation} examines estimation, initialization, and
adaptation on all four nonlinear tasks, together with the completed
layer-path interventions on KdV and KS. Appendix~\ref{app:ablation_results}
reports all five seeds, means, and sample standard deviations.

\begin{table}[!htbp]
\centering
\caption{\textbf{Contributions of estimation and correction.} Median GMean100 over five seeds. The fitted-only map is deterministic. Bold compares the four matched initialization/adaptation controls; Burgers is scaled by $10^{-3}$. Full seed statistics: Appendix~\ref{app:ablation_results}.}
\label{tab:main-ablation}
\begingroup\small
\setlength{\tabcolsep}{2pt}\renewcommand{\arraystretch}{1.05}
\begin{tabular*}{\linewidth}{@{\extracolsep{\fill}}lrrrr@{}}
\toprule
Setting & KdV & KS & Burgers & Kolmogorov \\
\midrule
Fitted map only & $0.841$ & $0.763$ & $523$ & $0.840$ \\
\addlinespace[2pt]\multicolumn{5}{l}{\textit{Matched controls: same SGNO corrector; direct-complex maps}} \\
Fixed identity & $0.0165$ & $0.0541$ & $3.63$ & $0.493$ \\
Trainable identity & $0.0143$ & $0.0498$ & $3.56$ & $0.445$ \\
Fixed fitted map & $0.0129$ & $0.0529$ & $3.68$ & $0.480$ \\
Trainable fitted & $\mathbf{0.0121}$ & $\mathbf{0.0477}$ & $\mathbf{3.38}$ & $\mathbf{0.427}$ \\
\midrule
\multicolumn{5}{l}{\textit{Original SGNO parameterization and layer-path interventions}} \\
SGNO & $0.0122$ & $0.0512$ & $3.34$ & $0.430$ \\
No spectral source & $173$ & $0.241$ & --- & --- \\
No affine path & $0.0105$ & $0.0640$ & --- & --- \\
\bottomrule
\end{tabular*}\endgroup
\par\smallskip\begin{minipage}{\linewidth}\footnotesize
Trainable controls use unit-disk projection; original SGNO uses the radial-square parameterization. The removed affine path is the pointwise map within each correction layer. ---: not evaluated.
\end{minipage}
\end{table}

The fitted map alone has GMean100 of 0.841, 0.763, 0.523, and 0.840 on
KdV, KS, Burgers, and Kolmogorov. Adding the correction and jointly
training both branches reduces the medians to 0.0122, 0.0512, 0.00334,
and 0.430. This comparison measures their combined effect.

The four matched controls share the SGNO corrector and initial weights,
data order, loss, and training budget. The outer map starts from the
identity or the fitted estimate and is either fixed or trainable.
Trainable maps use direct-complex parameters with unit-disk projection.
A fitted start lowers the median relative to a trainable identity on all
four tasks, with improvements in 3/5, 2/5, 2/5, and 3/5 paired seeds.
Allowing the fitted map to adapt also lowers all four medians, with
improvements in 4/5, 4/5, 3/5, and 5/5 seeds. The median improvements
therefore coexist with run-dependent effects, especially for initialization.

The path controls retain the original SGNO parameterization and retrain
with one layer path removed. Removing the nonlinear spectral source
raises median GMean100 from 0.0122 to 173 on KdV and from 0.0512 to
0.241 on KS. Removing the pointwise affine map gives 0.0105 and 0.0640,
respectively, showing task-dependent effects. This local map is distinct
from the outer fitted estimate. These interventions test path
contributions while retaining the shared generator construction.
Removing the output filter and zero-input subtraction changes the four
task medians by at most 3.8\%
(Appendix~\ref{app:output_processing}).

\subsection{Long-horizon prediction}
\label{subsec:tail_behavior}
We evaluate recursive prediction over 200 steps, four times the span
of each training trajectory. Across all five training runs and test
initial states, the median geometric-mean error over steps 101--200
(GMean101--200) is 0.0418 for SGNO versus 0.108 for IS-FNO(1) on KdV,
0.491 versus 0.715 on KS, and 0.00797 versus 0.0155 on Burgers.
Table~\ref{tab:temporal-compact} reports the complete late-window and
endpoint statistics, including between-run variation.

To characterize error growth within the same late window, we fit a
straight line to each run's trajectory-averaged nRMSE over steps
101--200. The slope measures the absolute change in nRMSE per prediction
step. Median slopes for SGNO versus IS-FNO(1) are
$0.318$ versus $0.575$ on KdV, $6.01$ versus $7.06$ on KS, and
$0.0615$ versus $0.104$ on Burgers, all in units of $10^{-3}$ per step.
SGNO has a lower slope in four of five paired runs on KdV and all five
on KS and Burgers. Appendix~\ref{app:current_temporal} provides the
definition and complete run statistics.

\begin{figure}[!htbp]
\centering
\includegraphics[width=\linewidth]{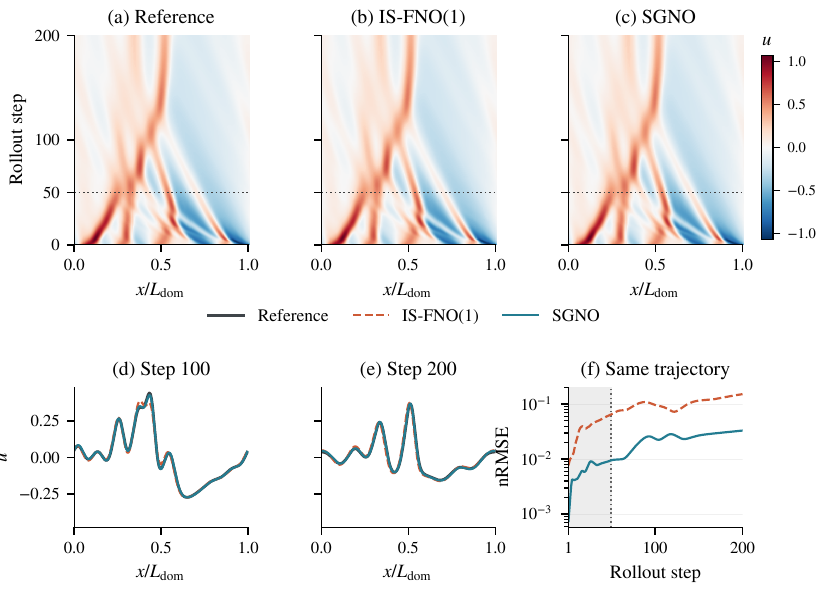}
\caption{\textbf{Qualitative comparison over a 200-step KdV rollout.}
(a--c) Reference, IS-FNO(1), and SGNO on a common color scale.
(d--e) Profiles at steps 100 and 200. (f) The same trajectory's nRMSE.
Dotted lines and gray shading mark the 50-transition training span;
prediction is recursive from step 1.
Complete statistics: Table~\ref{tab:temporal-compact}.}
\label{fig:rollout-curves}
\end{figure}

Figure~\ref{fig:rollout-curves} gives a qualitative KdV comparison.
Predictions are fed back without reference-state correction. The
space--time fields show the evolving wave structure; profiles at steps
100 and 200 compare its position and shape. The accompanying nRMSE curve
follows the same trajectory throughout the rollout. For this example,
the endpoint error is 0.0332 for SGNO and 0.153 for IS-FNO(1).
Appendix~\ref{app:volume_analysis} gives complementary volume and slice
views of the 2D example in Figure~\ref{fig:intro-motivation}(c).

\subsection{Amplitude and phase accuracy}
\label{subsec:propagation_mechanism}
The phase example in Section~\ref{subsec:learning_evolution} motivates
examining whether improved prediction accuracy is accompanied by lower
amplitude and phase-related errors. We examine the nonlinear KS task
after 200 recursive predictions. Equation~\eqref{eq:phase-magnitude-split}
separates squared Fourier coefficient error into nonnegative amplitude
and amplitude-weighted phase terms.

\begin{figure}[!htbp]
\centering
\includegraphics[width=\linewidth]{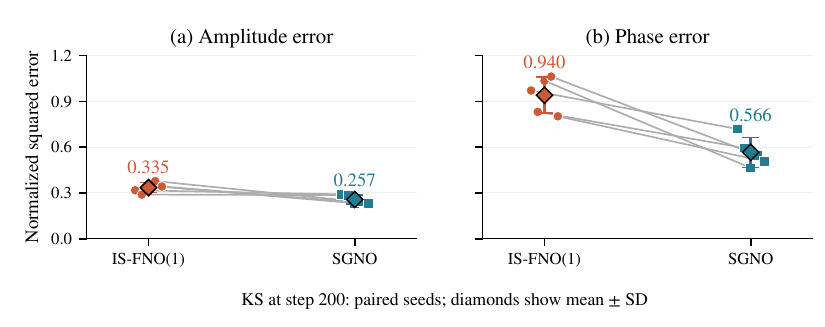}
\caption{\textbf{Amplitude and phase errors on KS at step 200.}
(a) Amplitude and (b) amplitude-weighted phase contributions to normalized
squared field error. Each point averages 30 test initial states for one
training run; gray lines connect paired runs. Diamonds and error bars
show the mean and sample standard deviation across five runs.}
\label{fig:mechanism-summary}
\end{figure}

Figure~\ref{fig:mechanism-summary} compares both components for
IS-FNO(1) and SGNO. Each component is normalized by reference energy
and averaged over 30 test initial states within each training run.
Across five runs, the mean amplitude error decreases from 0.335 to
0.257, while the mean phase contribution decreases from 0.940 to 0.566.
Both components are lower for SGNO in each paired run. On KS, the
lower endpoint error therefore reflects improvements in both amplitude
and phase-related field accuracy.

Field norms and correlation provide a complementary view of the same
KS endpoint. We measure the predicted-to-reference squared-norm ratio,
whose ideal value is one, and the normalized field inner product,
whose ideal value is also one. After averaging over test initial states
within each run, the five-run median squared-norm ratio is 1.05 for
SGNO and 1.14 for IS-FNO(1); the corresponding field correlations are
0.601 and 0.402. SGNO thus more closely matches the reference field's
overall magnitude and spatial structure at step 200.

These diagnostics describe errors in the complete predicted field,
while the interventions in Section~\ref{subsec:attribution} evaluate
individual design choices. Appendix~\ref{app:spectral_diagnostics}
gives the definitions and aggregation procedures.

\section{Discussion and Conclusion}
\label{sec:limitations}\label{sec:conclusion}
Long-horizon accuracy depends on both how existing errors propagate and
how accurately each prediction describes the next state. The phase
example makes this distinction concrete: preserving amplitude can still
leave a persistent error in motion. SGNO uses adjacent training states to
estimate a constrained Fourier map, then learns the changes it does not
describe. Joint training coordinates this estimate with a state-dependent
correction whose layers link the weights of existing and new features.
The linear tasks test recovery of the numerical update, while the
nonlinear tasks evaluate the complete estimate--correction predictor.
The component studies distinguish initialization, joint adaptation, and
the correction-layer paths. Across the four nonlinear tasks, a fitted
start and joint adaptation both lower the median within the matched
controls, with effects that vary across training runs. SGNO also achieves
lower median late-window errors on the three 1D nonlinear tasks.
The KS diagnostic shows lower amplitude and phase-related errors
at step 200, complementing the trajectory-level comparisons.
The present evaluation covers periodic scalar fields at fixed grids,
parameters, means, and output intervals. Varying these conditions,
extending to nonperiodic boundaries and nonlinear 3D systems, and deriving
quantitative rollout guarantees are natural next directions.

\clearpage
\subsection*{Reproducibility statement}
Appendices~\ref{app:method_details}--\ref{app:results} provide derivations,
implementation, data and training settings, and supporting statistics.
Code, configurations, and reproduction commands are available at
\url{https://github.com/cruiseresearchgroup/SGNO}.

\subsection*{AI use statement}
Generative AI tools assisted with language editing.

\bibliography{references}
\bibliographystyle{iclr2027_conference}

\clearpage
\appendix
\renewcommand{\thesection}{A\arabic{section}}
\raggedbottom
\renewcommand{\ms}[2]{\ensuremath{#1\,\pm\,#2}}
\noindent\textbf{Appendix organization.}
Appendix~\ref{app:method_details} develops the method derivations;
Appendix~\ref{app:protocol} specifies data, models, and evaluation.
Appendices~\ref{app:full_comparison}, \ref{app:ablation_results}, and
\ref{app:current_temporal} expand the main comparison, component controls,
and long-horizon results, respectively. Appendix~\ref{app:spectral_diagnostics}
defines the amplitude and phase error decomposition; Appendix~\ref{app:volume_analysis}
gives the additional 2D visualization.
\medskip
\section{Method extensions}
\label{app:method_details}\label{app:proofs}
\subsection{Phase-error accumulation}
\label{app:phase_accumulation_derivation}
For a nonzero complex coefficient $z$, fixed real phase increment $\omega$,
and increment error $\varepsilon$, the two iterates after $n$ steps are
$z_n=e^{in\omega}z$ and $\widetilde z_n=e^{in(\omega+\varepsilon)}z$.
Euler's identity gives
\begin{equation}
\frac{|\widetilde z_n-z_n|}{|z|}
=|e^{in\varepsilon}-1|
=\left|2i e^{in\varepsilon/2}\sin(n\varepsilon/2)\right|
=2|\sin(n\varepsilon/2)|.
\end{equation}
Expanding sine at zero yields
$n|\varepsilon|+\mathcal O((n|\varepsilon|)^3)$, and relative squared error
$n^2\varepsilon^2+\mathcal O((n|\varepsilon|)^4)$.
The condition is small \emph{accumulated} phase error $n|\varepsilon|\ll1$.
For arbitrary $n$, the exact relative error is bounded by two and can oscillate.

\subsection{Constrained affine regression}
\label{app:propagation_construction}\label{sec:theory}
For the paired Fourier coefficients $x_j,y_j$ in
Section~\ref{subsec:propagator_fit}, let $\bar x,\bar y$ be sample means
and define $s_{xx}=\sum_j|x_j-\bar x|^2$ and
$s_{xy}=\sum_j\overline{(x_j-\bar x)}(y_j-\bar y)$; the outer overline
denotes complex conjugation. The projection onto the unit disk
$\mathbb D=\{z\in\mathbb C:|z|\leq1\}$ is
$p_{\mathbb D}(z)=z/\max\{1,|z|\}$.
\begin{proposition}[Affine propagation fit]
\label{prop:affine_fit}
For $\mathbf k\in\mathcal K$ and $s_{xx}>0$, the unique solution of
Equation~\eqref{eq:affine_fit} is
\begin{equation}
m_{\mathbf k}^0=p_{\mathbb D}(s_{xy}/s_{xx}),\qquad
c_{\mathbf k}^0=\bar y-m_{\mathbf k}^0\bar x.
\label{eq:affine_solution}
\end{equation}
The residuals $r_j=y_j-m_{\mathbf k}^0x_j-c_{\mathbf k}^0$ satisfy
$\sum_jr_j=0$. If the gain constraint is inactive, they also satisfy
$\sum_j\overline{(x_j-\bar x)}r_j=0$.
\end{proposition}
\begin{proof}
For fixed $m$, minimizing over the real and imaginary parts of $c$ gives
$c=\bar y-m\bar x$. Completing the square then gives
\begin{equation}
\sum_j|(y_j-\bar y)-m(x_j-\bar x)|^2
=\kappa+s_{xx}|m-s_{xy}/s_{xx}|^2,
\end{equation}
where $\kappa$ is independent of $m$. Since $s_{xx}>0$, the unique
disk-constrained minimizer is the Euclidean projection of $s_{xy}/s_{xx}$.
Substituting the fitted intercept proves the zero residual sum. Interior
optimality gives the stated orthogonality. On the boundary,
$\sum_j\overline{(x_j-\bar x)}r_j$ is a nonnegative real multiple of
$m_{\mathbf k}^0$, the normal-cone optimality condition.
\end{proof}
Outside $\mathcal K$, the offset is zero, and the same argument uses
$s_{xx}=\sum_j|x_j|^2$ and $s_{xy}=\sum_j\overline{x_j}y_j$.
The fit gives a constrained affine approximation to the observed
discrete update, including on nonlinear tasks.
The jointly trained multiplier is
$m_{\mathbf k}=\exp(-\rho_{\mathbf k}^2+i\omega_{\mathbf k})$.
At $\rho_{\mathbf k}=0$, its radial derivative is zero, motivating the
direct-complex controls in Appendix~\ref{app:component_protocol}.
The gain constraint controls the linear multiplier; the complete predictor
also includes the learned correction.

\paragraph{Affine structure and spatial translations.}
\label{app:affine_equivariance}
For a spatial translation matrix $\mathbf T$, a Fourier-diagonal linear
map $\mathbf L$, and an offset vector $\mathbf b$, the affine map
$a(\mathbf q)=\mathbf L\mathbf q+\mathbf b$ obeys
$a(\mathbf T\mathbf q)-\mathbf T a(\mathbf q)=\mathbf b-\mathbf T\mathbf b$.
Exact translation equivariance therefore requires a spatially constant
offset. The retained nonzero-mode offsets and coordinate inputs permit
position dependence, as needed for spatially forced Kolmogorov flow.

\subsection{Derivation of the correction weights}
\label{app:feature_update_derivation}
We derive the two weights in Equation~\eqref{eq:main_corrector_layer}
at a fixed layer $\ell$ and frequency $\mathbf k$, suppressing the
frequency argument of $\boldsymbol\Lambda_\ell$. Let
$\mathbf z(\tau)\in\mathbb C^C$ be an auxiliary channel vector,
$\tau\in[0,1]$ its layer coordinate, and $\mathbf g\in\mathbb C^C$ a
source held constant over this interval. Consider
\begin{equation}
 \frac{d\mathbf z}{d\tau}=\boldsymbol\Lambda_\ell\mathbf z+\mathbf g.
 \label{eq:auxiliary_feature_equation}
\end{equation}
Multiplying by $e^{-\tau\boldsymbol\Lambda_\ell}$ gives
\begin{equation}
 \frac{d}{d\tau}\!\left(e^{-\tau\boldsymbol\Lambda_\ell}\mathbf z(\tau)\right)
 =e^{-\tau\boldsymbol\Lambda_\ell}\mathbf g.
\end{equation}
Integrating from zero to one and multiplying by
$e^{\boldsymbol\Lambda_\ell}$ yields
\begin{equation}
\mathbf z(1)=e^{\boldsymbol\Lambda_\ell}\mathbf z(0)
+\int_0^1 e^{(1-\tau)\boldsymbol\Lambda_\ell}\mathbf g\,d\tau
=e^{\boldsymbol\Lambda_\ell}\mathbf z(0)
+\phi_1(\boldsymbol\Lambda_\ell)\mathbf g.
\label{eq:feature_update_motivation}
\end{equation}
To evaluate the integral, apply it to each diagonal entry $\lambda$ of
$\boldsymbol\Lambda_\ell$:
\begin{equation}
 \int_0^1 e^{(1-\tau)\lambda}\,d\tau
 =\begin{cases}(e^\lambda-1)/\lambda,&\lambda\ne0,\\1,&\lambda=0,\end{cases}
 =\phi_1(\lambda).
\end{equation}
For $\lambda\leq0$, the integrand is positive and at most one, giving
$0<\phi_1(\lambda)\leq1$; likewise $0<e^\lambda\leq1$.
Both weights approach one as $\lambda\to0$, so the update approaches
$\mathbf z(0)+\mathbf g$.

The Fourier update in the correction layer uses
\begin{equation}
 \mathbf z(0)=[f_r(\mathbf v_\ell)]_{\mathbf k},\qquad
 \mathbf g=\mathbf R_\ell(\mathbf k)\mathbf A_\ell(\mathbf k)
 [f_r(s_\ell(\mathbf v_\ell))]_{\mathbf k}.
\end{equation}
Thus one generator determines both the weight on existing features and
the weight on the source computed from the layer input. The source is
held fixed only within this auxiliary step and is recomputed at each layer.
The coordinate $\tau$ describes feature processing rather than physical
time; each next-state prediction evaluates all $L$ layers.

\paragraph{Meaning of generator sharing.}
At each retained mode, define the effective channel-mixing matrix
$\widetilde{\mathbf A}_\ell=\phi_1(\boldsymbol\Lambda_\ell)\mathbf A_\ell$.
Since $\mathbf R_\ell$ and $\phi_1(\boldsymbol\Lambda_\ell)$ are diagonal,
\begin{equation}
\phi_1(\boldsymbol\Lambda_\ell)\mathbf R_\ell\mathbf A_\ell
=\mathbf R_\ell\widetilde{\mathbf A}_\ell,
\qquad \|\widetilde{\mathbf A}_\ell\|_2\leq\|\mathbf A_\ell\|_2\leq1.
\label{eq:shared_effective_mixer}
\end{equation}
Here the matrix norm is the spectral norm; the inequality follows from
$0<\phi_1(\lambda)\leq1$. Sharing couples the source weight to the
input-feature weight; the source factor can be absorbed into a mixing
matrix with the same norm bound. Sharing thus constrains the
parameterization within this norm-bounded source-map family by linking
the two feature-update weights.

\subsection{Fourier representation and correction implementation}
\label{app:parameterization}\label{app:real_fft}
For $C$ channels, the real FFT maps $\mathbb R^{n_gC}$ to a real subspace
of $\mathbb C^{n_fC}$, where the number of stored Fourier modes is
$n_f=(\prod_{j<d}N_j)(\lfloor N_d/2\rfloor+1)$.
Parseval weights are one at last-axis frequency zero and the even-grid
Nyquist frequency, and two elsewhere. With unitary normalization, the
transform is an isometry for this weighted real inner product.
On the zero and Nyquist planes, let
$\mathbf k^\star=(-k_1,\ldots,-k_{d-1},k_d)$ modulo grid sizes denote the
conjugate partner of $\mathbf k$. For a Fourier vector $\mathbf z$, let
$z_{\mathbf k,c}$ be its coefficient at mode $\mathbf k$ and channel
$c\in\{1,\ldots,C\}$. The real reconstruction projection is
\begin{equation}
[p_H(\mathbf z)]_{\mathbf k,c}
=\tfrac12\left(z_{\mathbf k,c}+\overline{z_{\mathbf k^\star,c}}\right),
\label{eq:hermitian_projection}
\end{equation}
and is the identity on other stored modes. Self-conjugate coefficients
become real. The inverse real FFT implements this projection implicitly.

Let $p_{\boldsymbol\theta}$ denote the pointwise feature lift and
$q_{\boldsymbol\theta}$ the pointwise readout at the start of Project.
They produce $\mathbf v_0=p_{\boldsymbol\theta}([\mathbf q,\mathbf x])$
and the raw scalar field
$d_{\boldsymbol\theta}(\mathbf q)=q_{\boldsymbol\theta}(\mathbf v_L)$.
Layers follow Equation~\eqref{eq:main_corrector_layer}, with untied
parameters. The mask and $p_H$ are grouped into irFFT in the diagram.
\paragraph{Generator and source weights.}

For the dimension-specific radial coordinate $\chi(\mathbf{k})$ defined below, the generator and channel scale are parameterized by
\begin{equation}
\begin{aligned}
\lambda_{\ell,c}(\mathbf{k})
&=-\operatorname{softplus}(a_{\ell,c}(\mathbf{k}))
 -\operatorname{softplus}\!\left(
b_{\ell,c}-\operatorname{softplus}(s_{\ell,c})\chi(\mathbf{k})\right),\\
r_{\ell,c}(\mathbf{k})
&=2\operatorname{sigmoid}\!\left(
q_{\ell,c}-\operatorname{softplus}(p_{\ell,c})\chi(\mathbf{k})\right).
\end{aligned}
\label{eq:parameterization}
\end{equation}
Here $a_{\ell,c}(\mathbf k)$, $b_{\ell,c}$, $s_{\ell,c}$,
$q_{\ell,c}$, and $p_{\ell,c}$ are learned real parameters;
only $a_{\ell,c}(\mathbf k)$ varies with the mode.
Here $\operatorname{softplus}(z)=\log(1+e^z)$ and
$\operatorname{sigmoid}(z)=(1+e^{-z})^{-1}$.
The values $\lambda_{\ell,c}(\mathbf{k})$ and $r_{\ell,c}(\mathbf{k})$
are the diagonal entries of $\boldsymbol{\Lambda}_\ell(\mathbf{k})$ and
$\mathbf{R}_\ell(\mathbf{k})$. These give $\lambda_{\ell,c}(\mathbf{k})<0$
and $0<r_{\ell,c}(\mathbf{k})<2$ for finite parameters in exact arithmetic.
The radius-dependent softplus term need not increase with frequency;
the parameterization does not impose monotone high-frequency damping.

At each retained mode, the effective mixing matrix is the raw matrix divided by the larger of one and its largest singular value. The two pointwise source maps use spectral normalization~\citep{miyato2018spectral}. The two-dimensional raw complex mixer is initialized as $C^{-2}(\boldsymbol{\Xi}+i\boldsymbol{\Upsilon})$, with independent entries of $\boldsymbol{\Xi},\boldsymbol{\Upsilon}$ drawn uniformly from $[0,1)$.

Write $m_{\mathrm{cut}}$ for the retained mode setting. The affine intercept
and corrector use the following bands on the stored real-FFT grid:
\begin{equation}
\begin{aligned}
\mathcal K&=\{\mathbf{k}:-m_{\mathrm{cut}}<k_j<m_{\mathrm{cut}}\ (j<d),\quad
           0\leq k_d<m_{\mathrm{cut}}\},\\
\mathcal K_{\mathrm{corr}}&=\{\mathbf{k}:-m_{\mathrm{cut}}\leq k_j<m_{\mathrm{cut}}\ (j<d),\quad
           0\leq k_d<m_{\mathrm{cut}}\}.
\end{aligned}
\label{eq:implemented_bands}
\end{equation}
Thus the corrector includes the negative endpoint in each full Fourier
axis. The mask in Equation~\eqref{eq:main_corrector_layer} has
$\mu_{\mathbf k}=1$ on $\mathcal K_{\mathrm{corr}}$ and zero elsewhere.
In two dimensions the radial coordinate uses signed frequencies,
\begin{equation}
 \chi(\mathbf{k})=\frac{\sqrt{k_1^2+k_2^2+10^{-12}}}
 {\sqrt{m_{\mathrm{cut}}^2+(m_{\mathrm{cut}}-1)^2}}.
\end{equation}
In one and three dimensions it uses the indices within each retained block,
\begin{equation}
 \chi(\mathbf{k})=\frac1{\sqrt d}\sqrt{
 \sum_{j=1}^d\left(\frac{i_j}{m_{\mathrm{cut}}-1}\right)^2+10^{-12}},
 \qquad
 i_j=\begin{cases}
 k_j,&k_j\geq0,\\
 k_j+m_{\mathrm{cut}},&k_j<0.
 \end{cases}
\end{equation}
In particular, $i_d=k_d$. The three-dimensional coordinate is a block-local
weighting coordinate rather than the physical frequency radius.

At each grid point, the source maps a channel vector $\mathbf{z}\in\mathbb R^C$ through hidden width $2C$:
\begin{equation}
 s_\ell(\mathbf{z})=\mathbf{U}_{\ell,2}\operatorname{gelu}(\mathbf{U}_{\ell,1}\mathbf{z}+\mathbf{b}_{\ell,1})+\mathbf{b}_{\ell,2},
\end{equation}
The weight matrices are $\mathbf{U}_{\ell,1}\in\mathbb R^{2C\times C}$ and
$\mathbf{U}_{\ell,2}\in\mathbb R^{C\times2C}$,
with bias vectors $\mathbf{b}_{\ell,1}\in\mathbb R^{2C}$ and
$\mathbf{b}_{\ell,2}\in\mathbb R^C$.
The same function is applied independently at every grid point.

For $|z|<10^{-4}$, the implementation evaluates
\begin{equation}
\phi_1(z)\approx
1+\frac z2+\frac{z^2}{6}+\frac{z^3}{24}
+\frac{z^4}{120}+\frac{z^5}{720}.
\label{eq:phi_taylor}
\end{equation}
Otherwise it uses $\operatorname{expm1}(z)/z$, where
$\operatorname{expm1}(z)=e^z-1$. This avoids division by a small number
while retaining a finite derivative at zero. The formulas use the exact
$\phi_1$; this expansion specifies its numerical evaluation.

\paragraph{Complete correction.}
Let $\mathbf{P}_{\mathrm{hi}}$ project onto the complement of the intercept band $\mathcal K$.
Let $\alpha$ be a learned scalar weight and
$\mathbf b=(b_0,\ldots,b_d)$ the real parameter vector of an output filter
$\mathbf G_{\mathbf b}\in\mathbb R^{n_g\times n_g}$.
The state-dependent contribution in Equation~\eqref{eq:physical_decomposition}
is
\begin{equation}
 r_{\boldsymbol{\theta}}(\mathbf{q})=\alpha\mathbf{G}_{\mathbf{b}}\!\left(
 d_{\boldsymbol{\theta}}(\mathbf{q})-\mathbf{P}_{\mathrm{hi}}d_{\boldsymbol{\theta}}(\mathbf{0})\right),
 \label{eq:sgno_full_map}
\end{equation}
where $\mathbf G_{\mathbf b}$ has gain at most one and the
subtraction removes the high-frequency response produced at zero input.
For signed integer frequencies $k_j$, its multiplier is
\begin{equation}
 g_{\mathbf{b}}(\mathbf{k})=\exp\!\left[-b_0^2-\sum_{j=1}^d b_j^2(2k_j/N_j)^2\right].
\end{equation}
The filter parameters are initialized with
$b_0=0.01$ and $b_j=0.1$ for $j\geq1$. The unconstrained scalar $\alpha$
starts at one; all these parameters are trained jointly. The final readout
weights and bias of $d_{\boldsymbol{\theta}}$ start at zero. The high-frequency
projection uses $\max_j|k_j|\geq m_{\mathrm{cut}}$, where
$m_{\mathrm{cut}}$ is the retained mode count.

The training-pair regression initializes $\mathbf m_{\boldsymbol{\eta}}$ on the full stored Fourier
grid and the intercept on $\mathcal K$ in~\eqref{eq:implemented_bands}. Inside that band,
it uses the centered statistics of Proposition~\ref{prop:affine_fit}; outside,
it uses $\sum_j|x_j|^2$ and $\sum_j\overline{x_j}y_j$ with zero intercept.
Real and complex statistics are accumulated in float64 and complex128, respectively. If a mode's denominator is at most
$10^{-12}$ times the largest denominator (with the smallest positive float64
normal value as a floor), its multiplier is initialized to one. After
unit-disk projection, magnitudes are floored at $10^{-12}$ to obtain a finite
logarithmic parameterization, and parameters are stored in complex64.

In Figure~\ref{fig:sgno_architecture}(a), Project groups this readout,
zero-input subtraction, output filter, scalar weight, and mean removal.
The upper branch's Restore mean returns
$\mathbf P_0\mathbf w+\mathbf Qa_{\boldsymbol\eta}(\mathbf q)$; their sum
is Equation~\eqref{eq:physical_decomposition}.

\FloatBarrier
\section{Reproduction details}
\label{app:protocol}
\subsection{Data and task configurations}
\label{app:data_configuration}
All tasks use periodic scalar fields and fixed parameters. Training contains
50 trajectories of 50 transitions; testing contains 200 transitions from
30 initial states in 1D/2D and ten in 3D. Table~\ref{tab:luo-task-settings}
specifies the observed grids and numerical updates.
\begin{table}[!htbp]
\centering
\caption{Task configurations and observed grids. A starred interval uses the
normalized output-step unit of an APEBench difficulty scenario; the other
intervals are in physical time. The domain side length is 1 except for
Kolmogorov ($2\pi$).}
\label{tab:luo-task-settings}
\footnotesize
\setlength{\tabcolsep}{3pt}
\begin{tabularx}{\linewidth}{l c c X}
\toprule
Task & Grid & $\Delta t$ & Configuration \\
\midrule
\multicolumn{4}{l}{\textbf{Linear tasks}} \\
Advection & $160$ & $1^\ast$ & $\boldsymbol{\gamma}=(0,-4,0,0,0)$ \\
Diffusion & $160$ & $1^\ast$ & $\boldsymbol{\gamma}=(0,0,4,0,0)$ \\
Dispersion & $160$ & $1^\ast$ & $\boldsymbol{\gamma}=(0,0,0,4,0)$ \\
Mixed dispersion & $64^2$ & $0.001$ & $u_t=0.00025(\partial_x+\partial_y)\Delta u$ \\
Axis diffusion & $64^2$ & $0.1$ & $\mathbf{D}=\operatorname{diag}(0.001,0.002)$ \\
Rotated diffusion & $64^2$ & $0.1$ & $D_{11}=D_{22}=0.0015$; $D_{12}=D_{21}=-0.0005$ \\
Diagonal diffusion & $32^3$ & $0.1$ & $\mathbf{D}=\operatorname{diag}(0.001,0.002,0.0004)$ \\
Unbalanced advection & $32^3$ & $0.1$ & $u_t=-\mathbf{v}\cdot\nabla u$; $\mathbf{v}=(0.01,-0.04,0.005)$ \\
Advection--diffusion & $32^3$ & $1^\ast$ & $\boldsymbol{\gamma}=(0,-4,4,0,0)$ \\
\midrule
\multicolumn{4}{l}{\textbf{Nonlinear tasks}} \\
KdV & $160$ & $1^\ast$ & $\boldsymbol{\gamma}=(0,0,0,-14,-9)$; $\boldsymbol{\delta}=(0,-2,0)$ \\
KS & $160$ & $1^\ast$ & $\boldsymbol{\gamma}=(0,0,-1.2,0,-15)$; $\boldsymbol{\delta}=(0,0,-6)$ \\
Burgers & $160$ & $0.1$ & $(c,a_2,a_4)=(-0.125,0.0012,0)$ \\
Kolmogorov & $64^2$ & $0.1$ & Viscosity $0.01$; drag $-0.1$; forcing mode 4, scale 1 \\
\bottomrule
\end{tabularx}
\end{table}

Here $u$ is the scalar field and $u_t$ its physical-time derivative;
bracketed $[t]$ elsewhere indexes stored states.
$D_{ij}$ denotes an entry of the diffusion matrix $\mathbf{D}$.
Here $\boldsymbol{\gamma}$ and $\boldsymbol{\delta}$ are APEBench difficulty coefficients, rather than
physical derivative coefficients. For $N$ points per axis in $d$ dimensions,
the normalized linear coefficients are $\alpha_0=\gamma_0$ and
$\alpha_j=\gamma_j/(N^j2^{j-1}d)$ for $j\geq1$; $\alpha_j$ multiplies
$\sum_{i=1}^d\partial_{x_i}^j u$ for $j\geq1$. With the scenario's unit
amplitude scale, the nonlinear coefficients are
$\boldsymbol{\beta}=(\delta_0,\delta_1/(Nd),\delta_2/(N^2d))$, multiplying
$u^2$, $\frac12\sum_i\partial_{x_i}(u^2)$, and
$\frac12(\|\nabla u\|^2-\langle\|\nabla u\|^2\rangle_{\mathbf{x}})$, respectively; $\langle\cdot\rangle_{\mathbf{x}}$ denotes the spatial mean. Thus the reported KdV configuration
includes fourth-order dissipation. For Burgers,
$u_t=(c/2)\partial_x(u^2)+a_2\partial_{xx}u+a_4\partial_{xxxx}u$;
the diffusion matrix $\mathbf{D}$ defines $u_t=\nabla\cdot(\mathbf{D}\nabla u)$.

Reference integration uses APEBench 0.1.1 and Exponax 0.1.0 Fourier
solvers~\citep{koehler2024apebench}. The linear tasks use their exact
semidiscrete Fourier update. KdV and KS use second-order exponential time
differencing with one reference step per output; Kolmogorov uses the same
order with 20 substeps~\citep{cox2002etd,kassam2005etdrk4}.
For KS, Exponax's \texttt{zero\_mode\_fix=True} removes the spatial mean of
the squared-gradient term, giving the mean-subtracted term above and
preserving the state mean. Burgers uses fourth-order integration
with ten substeps on 640 points, followed by aligned sampling on 160 points.
This reference is integrated in float64 and stored in float32;
the remaining nonlinear references use float32. Nonlinear evaluation uses
the $2/3$ dealiasing setting.
For Burgers, each training pair consists of two sampled fine-grid states.
Subsampling need not yield an autonomous deterministic update on the
observed grid; the model learns a one-step predictor from those observations.

Initial fields use APEBench's truncated Fourier generator with cutoff 5,
zero mean, and maximum absolute value normalized to one. KS and Kolmogorov undergo 500 reference output steps of warm-up before the stored
initial state; all other tasks use no warm-up. Training and test data use separate fixed data-generation seeds,
distinct from the model-training seeds. Evaluation uses the first 200 transitions of each
stored test trajectory.

\paragraph{Numerical target and mean convention.}
\label{app:mean_scope}
Evaluation uses the specified numerical reference and observation protocol.
For example, a 512-substep float64 KdV reference differs from the benchmark
reference by mean nRMSE 0.470 at step 50 and 0.938 at step 200; benchmark
accuracy therefore measures emulation of that update, rather than of a
temporally converged PDE solution. For the same KdV coefficients, a
separate reference check on independent initial states compares
512 and 1024 substeps, with maximum validation nRMSE 0.00720. This quantifies the effect of
temporal refinement at the fixed spatial grid. All stored states have approximately
zero spatial mean (maximum absolute mean $2.7243\times10^{-7}$).
For a scalar $b$ and the all-ones vector $\mathbf1$,
Equation~\eqref{eq:physical_decomposition} gives
$f_{\boldsymbol\theta}(\mathbf w+b\mathbf1)
=f_{\boldsymbol\theta}(\mathbf w)+b\mathbf1$.
Evaluation is restricted to the fixed-mean setting, since changing the
conserved mean can change nonlinear transport.

\subsection{Models and optimization}
\label{app:model_configuration}\label{app:correction_implementation}
SGNO uses width/modes/layers $(11,26,7)$, $(5,10,9)$, and $(10,6,2)$
in 1D, 2D, and 3D. Its parameter-array element counts are
30,683, 58,708, and 210,966; counting complex entries as two real
degrees of freedom gives 52,812, 106,010, and 401,900.
An SGNO layer uses two forward FFTs and one inverse FFT.

The affine map is fitted to training pairs and the final correction head
starts at zero. Adam uses batch size 20, $(\beta_1,\beta_2)=(0.9,0.999)$,
$\epsilon=10^{-8}$, zero weight decay, and no clipping. The learning rate
warms from zero to $10^{-3}$ over 2,000 updates and then decays by a cosine
schedule to update 10,000. All adjacent pairs are reshuffled each epoch.
The fixed final checkpoint is evaluated for each of the five training runs. The common
2D initialization was chosen using separate development trajectories
through step 200 and frozen before the independent test set.

FNO and CNO use the common one-step MSE protocol. Their 1D/2D parameter
counts are 53,062/102,713 and 56,317/105,012, respectively.
KNO retains prediction--reconstruction training: widths are 30 for
advection/diffusion/dispersion/KdV/KS, 40 for Burgers, 17 for mixed
dispersion/Kolmogorov, 22 for the two 2D diffusion tasks, and 15 in 3D.
IS-FNO(1) retains the invertible body with width/modes $(15,16)$ in 1D
and $(6,10)$ in 2D (29,742/51,275 parameters), adapted to one-step state
MSE and physical-state feedback. No 3D IS-FNO model was evaluated.

\paragraph{Precision and aggregation.}
\label{app:current_comparison_precision}\label{app:metrics}
Evaluations of SGNO, IS-FNO, and the component controls use
float32/complex64 with TensorFloat-32 (TF32) disabled; other main baselines
retain their recorded evaluations. For seed $s$ and $n$ initial states,
\begin{equation}
\bar e_s^{[t]}=\frac1n\sum_{j=1}^{n}
\frac{\|\widetilde{\mathbf u}_{s,j}^{[t]}-\mathbf u_j^{[t]}\|_{\rm RMS}}
{\max(\|\mathbf u_j^{[t]}\|_{\rm RMS},10^{-12})},\qquad
G_{s,a:b}=\exp\!\left[\frac1{b-a+1}\sum_{t=a}^{b}
\log(\bar e_s^{[t]}+10^{-12})\right].
\label{eq:gmean100}
\end{equation}
Here $\|\mathbf v\|_{\rm RMS}=\|\mathbf v\|_2/\sqrt{n_g}$.
This is the 1D/2D convention; 3D retains the original $\ell_2$
denominator plus $10^{-12}$ and float32 subtraction before float64
norm accumulation. GMean100 uses $a=1,b=100$; the late window uses
$a=101,b=200$. Main Tables~\ref{tab:main-comparison} and
\ref{tab:main-ablation} use the median of the five seed-level GMean100
values, with the same five training seeds for every evaluated group.
Appendices~\ref{app:full_comparison} and~\ref{app:ablation_results} give
the individual scores, arithmetic mean, and sample SD ($\mathrm{ddof}=1$).
The median summarizes the middle run; the accompanying full statistics
describe the spread across training seeds.
A group containing a non-finite seed is marked
NF, without recomputing over surviving seeds. Dashes denote unevaluated
configurations, and large finite values are retained.

\subsection{Component-control protocols}
\label{app:component_protocol}\label{app:factorial_results}
On all four nonlinear tasks, the initialization control uses a fixed identity, a trainable identity
start, a fixed fitted map, or a trainable fitted start. The two trainable
maps use direct complex multipliers with post-update unit-disk projection,
so attenuation can change from identity. The corrector initialization,
data order, loss, output processing, and update budget are fixed. These
controls are compared within the direct-complex group, separately from
the original radial-square SGNO checkpoints.
The fitted-only map is a deterministic regression with no neural updates.
The layer-path controls remove either the nonlinear spectral source or
the pointwise affine map within each correction layer. Each variant is
retrained on KdV and KS using five training seeds with the same training protocol;
the outer fitted map and output processing are retained.

\subsection{Qualitative visualization protocols}
\label{app:phase_example}
Figure~\ref{fig:intro-motivation} uses an additional 2D KdV-type
configuration (\texttt{dispconv2d\_s1}), combining convection, dispersion,
and hyperdiffusion on a $64\times64$ grid, distinct from the 1D KdV
benchmark in Table~\ref{tab:main-comparison}.
It shows stored FP32 predictions for the same test initial state.
Both models receive 10,000 updates using trajectories spanning 50
transitions; IS-FNO uses its native 20-state relative-error objective.
Predictions start from the same initial state without reference resets.
The reference volume uses Exponax/vAPE4D rendering. At step 200,
both contour panels use the central region $[0.25,0.75]^2$ and common
levels $(-1,0,1)$ times the full reference spatial RMS, with native-grid
linear interpolation and no smoothing. The displayed relative $L^2$
errors are evaluated on this region of this trajectory.
On the full spatial domain of the same step-200 fields, relative
$L^2$ errors are 0.254 for IS-FNO(20) and 0.0546 for SGNO.
Figure~\ref{fig:rollout-curves} uses saved one-step-trained, full-FP32
predictions for the 1D KdV configuration. Both models start from the same
test initial state. The field panels share a color scale, the two profile
panels share axis limits, and the nRMSE curve uses that same trajectory.
Table~\ref{tab:temporal-compact} reports the complete five-run statistics
over all test initial states.

\FloatBarrier
\section{Detailed prediction comparisons}
\label{app:results}\label{app:controlled_evidence}

\subsection{Full main-comparison statistics}
\label{app:full_comparison}
Tables~\ref{tab:main-all-seeds} and~\ref{tab:main-nonlinear-seeds}
expand the linear and nonlinear entries of
Table~\ref{tab:main-comparison}, using the same fixed training seeds and
per-seed metric. The main table's medians are computed from these values
before rounding; the mean and sample SD summarize between-seed variation.
The missing 3D IS-FNO entries are unevaluated configurations.

\begingroup\small
\setlength{\tabcolsep}{2.4pt}
\renewcommand{\arraystretch}{1.15}
\setlength{\LTleft}{\fill}\setlength{\LTright}{\fill}
\setlength{\LTcapwidth}{\linewidth}
\begin{longtable}{@{}lrrrrrrrr@{}}
\caption{\textbf{Linear prediction: full seed statistics.} GMean100 for the linear block of Table~\ref{tab:main-comparison}. Summary columns use the same five seed scores shown on the right.}\label{tab:main-all-seeds}\\
\toprule
Method & \multicolumn{3}{c}{Summary} & \multicolumn{5}{c}{Training run} \\
\cmidrule(lr){2-4}\cmidrule(l){5-9}
 & Median & Mean & SD & A & B & C & D & E \\
\midrule
\endfirsthead
\multicolumn{9}{l}{\tablename\ \thetable\ (continued)}\\
\toprule
Method & \multicolumn{3}{c}{Summary} & \multicolumn{5}{c}{Training run} \\
\cmidrule(lr){2-4}\cmidrule(l){5-9}
 & Median & Mean & SD & A & B & C & D & E \\
\midrule
\endhead
\midrule\multicolumn{9}{r}{\footnotesize Continued on the next page}\\
\endfoot
\bottomrule
\endlastfoot
\addlinespace[5pt]\multicolumn{9}{l}{\textit{Advection (1D)}}\\*
FNO & $0.0236$ & $0.0244$ & $0.0026$ & $0.0271$ & $0.0222$ & $0.0236$ & $0.0219$ & $0.0270$ \\*
CNO & $0.854$ & $2.03\mathrm{e}{3}$ & $4.5\mathrm{e}{3}$ & $0.894$ & $0.727$ & $0.757$ & $1.01\mathrm{e}{4}$ & $0.854$ \\*
KNO & $0.0280$ & $0.0271$ & $0.0076$ & $0.0303$ & $0.0280$ & $0.0369$ & $0.0243$ & $0.0162$ \\*
IS-FNO(1) & $\mathrm{NF}_{1/5}$ & $\mathrm{NF}_{1/5}$ & $\mathrm{NF}_{1/5}$ & $\mathrm{NF}$ & $0.0353$ & $0.0117$ & $0.00494$ & $0.00703$ \\*
\textbf{SGNO} & $2.54\mathrm{e}{-6}$ & $2.52\mathrm{e}{-6}$ & $2.6\mathrm{e}{-7}$ & $2.92\mathrm{e}{-6}$ & $2.57\mathrm{e}{-6}$ & $2.37\mathrm{e}{-6}$ & $2.54\mathrm{e}{-6}$ & $2.22\mathrm{e}{-6}$ \\
\addlinespace[5pt]\multicolumn{9}{l}{\textit{Diffusion (1D)}}\\*
FNO & $0.0198$ & $0.0206$ & $0.0019$ & $0.0191$ & $0.0214$ & $0.0236$ & $0.0198$ & $0.0193$ \\*
CNO & $0.207$ & $1.83$ & $3.5$ & $0.207$ & $0.163$ & $8.01$ & $0.585$ & $0.168$ \\*
KNO & $0.0274$ & $0.239$ & $0.49$ & $0.0312$ & $0.0274$ & $1.11$ & $0.0196$ & $0.0105$ \\*
IS-FNO(1) & $1.64\mathrm{e}{-4}$ & $1.75\mathrm{e}{-4}$ & $6.3\mathrm{e}{-5}$ & $2.83\mathrm{e}{-4}$ & $1.64\mathrm{e}{-4}$ & $1.27\mathrm{e}{-4}$ & $1.67\mathrm{e}{-4}$ & $1.36\mathrm{e}{-4}$ \\*
\textbf{SGNO} & $8.12\mathrm{e}{-6}$ & $7.64\mathrm{e}{-6}$ & $9.5\mathrm{e}{-7}$ & $6.58\mathrm{e}{-6}$ & $8.56\mathrm{e}{-6}$ & $8.28\mathrm{e}{-6}$ & $6.64\mathrm{e}{-6}$ & $8.12\mathrm{e}{-6}$ \\
\addlinespace[5pt]\multicolumn{9}{l}{\textit{Dispersion (1D)}}\\*
FNO & $0.0322$ & $0.0337$ & $0.0051$ & $0.0329$ & $0.0291$ & $0.0322$ & $0.0425$ & $0.0320$ \\*
CNO & $0.265$ & $6.62\mathrm{e}{8}$ & $1.5\mathrm{e}{9}$ & $0.0674$ & $3.31\mathrm{e}{9}$ & $6.70$ & $0.0706$ & $0.265$ \\*
KNO & $0.0277$ & $0.0284$ & $0.0086$ & $0.0339$ & $0.0277$ & $0.0390$ & $0.0252$ & $0.0165$ \\*
IS-FNO(1) & $4.87\mathrm{e}{-5}$ & $5.20\mathrm{e}{-5}$ & $1.4\mathrm{e}{-5}$ & $5.65\mathrm{e}{-5}$ & $3.45\mathrm{e}{-5}$ & $4.87\mathrm{e}{-5}$ & $4.78\mathrm{e}{-5}$ & $7.26\mathrm{e}{-5}$ \\*
\textbf{SGNO} & $5.01\mathrm{e}{-6}$ & $5.06\mathrm{e}{-6}$ & $2.0\mathrm{e}{-7}$ & $5.41\mathrm{e}{-6}$ & $4.97\mathrm{e}{-6}$ & $5.03\mathrm{e}{-6}$ & $5.01\mathrm{e}{-6}$ & $4.90\mathrm{e}{-6}$ \\
\addlinespace[5pt]\multicolumn{9}{l}{\textit{Mixed dispersion (2D)}}\\*
FNO & $0.186$ & $0.171$ & $0.048$ & $0.224$ & $0.106$ & $0.200$ & $0.137$ & $0.186$ \\*
CNO & $0.0516$ & $0.0504$ & $0.0034$ & $0.0477$ & $0.0516$ & $0.0543$ & $0.0460$ & $0.0523$ \\*
KNO & $0.167$ & $0.172$ & $0.0075$ & $0.166$ & $0.166$ & $0.167$ & $0.181$ & $0.179$ \\*
IS-FNO(1) & $0.00122$ & $0.00136$ & $6.9\mathrm{e}{-4}$ & $0.00178$ & $0.00122$ & $5.65\mathrm{e}{-4}$ & $0.00230$ & $9.60\mathrm{e}{-4}$ \\*
\textbf{SGNO} & $6.06\mathrm{e}{-6}$ & $6.05\mathrm{e}{-6}$ & $1.9\mathrm{e}{-7}$ & $6.06\mathrm{e}{-6}$ & $6.07\mathrm{e}{-6}$ & $5.95\mathrm{e}{-6}$ & $5.83\mathrm{e}{-6}$ & $6.34\mathrm{e}{-6}$ \\
\addlinespace[5pt]\multicolumn{9}{l}{\textit{Axis diffusion (2D)}}\\*
FNO & $0.0804$ & $0.0733$ & $0.014$ & $0.0807$ & $0.0545$ & $0.0804$ & $0.0631$ & $0.0878$ \\*
CNO & $0.309$ & $0.311$ & $0.0076$ & $0.312$ & $0.308$ & $0.302$ & $0.309$ & $0.323$ \\*
KNO & $0.0172$ & $0.0194$ & $0.0045$ & $0.0162$ & $0.0204$ & $0.0161$ & $0.0172$ & $0.0269$ \\*
IS-FNO(1) & $0.00354$ & $0.0328$ & $0.060$ & $0.140$ & $0.00354$ & $0.00143$ & $0.0161$ & $0.00269$ \\*
\textbf{SGNO} & $7.74\mathrm{e}{-6}$ & $7.77\mathrm{e}{-6}$ & $4.6\mathrm{e}{-8}$ & $7.73\mathrm{e}{-6}$ & $7.74\mathrm{e}{-6}$ & $7.80\mathrm{e}{-6}$ & $7.83\mathrm{e}{-6}$ & $7.73\mathrm{e}{-6}$ \\
\addlinespace[5pt]\multicolumn{9}{l}{\textit{Rotated diffusion (2D)}}\\*
FNO & $0.0783$ & $0.0732$ & $0.016$ & $0.0802$ & $0.0536$ & $0.0783$ & $0.0610$ & $0.0927$ \\*
CNO & $0.319$ & $0.333$ & $0.057$ & $0.330$ & $0.293$ & $0.431$ & $0.293$ & $0.319$ \\*
KNO & $0.0461$ & $0.0465$ & $0.0026$ & $0.0461$ & $0.0433$ & $0.0448$ & $0.0486$ & $0.0496$ \\*
IS-FNO(1) & $0.0123$ & $0.0242$ & $0.037$ & $0.0890$ & $0.0158$ & $0.00139$ & $0.0123$ & $0.00251$ \\*
\textbf{SGNO} & $7.57\mathrm{e}{-6}$ & $7.58\mathrm{e}{-6}$ & $4.8\mathrm{e}{-8}$ & $7.57\mathrm{e}{-6}$ & $7.55\mathrm{e}{-6}$ & $7.54\mathrm{e}{-6}$ & $7.66\mathrm{e}{-6}$ & $7.58\mathrm{e}{-6}$ \\
\addlinespace[5pt]\multicolumn{9}{l}{\textit{Diagonal diffusion (3D)}}\\*
FNO & $0.166$ & $1.20$ & $2.3$ & $5.32$ & $0.160$ & $0.166$ & $0.150$ & $0.181$ \\*
CNO & $0.331$ & $0.330$ & $0.011$ & $0.340$ & $0.328$ & $0.313$ & $0.338$ & $0.331$ \\*
KNO & $0.0198$ & $0.0229$ & $0.0093$ & $0.0123$ & $0.0174$ & $0.0301$ & $0.0198$ & $0.0350$ \\*
\textbf{SGNO} & $5.80\mathrm{e}{-6}$ & $5.85\mathrm{e}{-6}$ & $1.6\mathrm{e}{-7}$ & $5.72\mathrm{e}{-6}$ & $5.74\mathrm{e}{-6}$ & $6.11\mathrm{e}{-6}$ & $5.80\mathrm{e}{-6}$ & $5.86\mathrm{e}{-6}$ \\
\addlinespace[5pt]\multicolumn{9}{l}{\textit{Unbalanced advection (3D)}}\\*
FNO & $0.784$ & $1.43$ & $1.1$ & $3.06$ & $0.722$ & $0.669$ & $0.784$ & $1.94$ \\*
CNO & $0.134$ & $0.135$ & $0.018$ & $0.161$ & $0.134$ & $0.131$ & $0.138$ & $0.110$ \\*
KNO & $0.0432$ & $0.0457$ & $0.013$ & $0.0311$ & $0.0450$ & $0.0432$ & $0.0431$ & $0.0661$ \\*
\textbf{SGNO} & $5.77\mathrm{e}{-6}$ & $5.77\mathrm{e}{-6}$ & $1.1\mathrm{e}{-7}$ & $5.93\mathrm{e}{-6}$ & $5.62\mathrm{e}{-6}$ & $5.77\mathrm{e}{-6}$ & $5.80\mathrm{e}{-6}$ & $5.72\mathrm{e}{-6}$ \\
\addlinespace[5pt]\multicolumn{9}{l}{\textit{Advection--diffusion (3D)}}\\*
FNO & $0.144$ & $0.138$ & $0.019$ & $0.110$ & $0.157$ & $0.144$ & $0.128$ & $0.150$ \\*
CNO & $1.44$ & $1.62$ & $0.47$ & $1.35$ & $1.81$ & $2.34$ & $1.15$ & $1.44$ \\*
KNO & $0.0171$ & $0.0166$ & $0.0039$ & $0.0123$ & $0.0220$ & $0.0135$ & $0.0183$ & $0.0171$ \\*
\textbf{SGNO} & $7.11\mathrm{e}{-6}$ & $7.11\mathrm{e}{-6}$ & $1.5\mathrm{e}{-7}$ & $7.11\mathrm{e}{-6}$ & $6.91\mathrm{e}{-6}$ & $7.01\mathrm{e}{-6}$ & $7.19\mathrm{e}{-6}$ & $7.31\mathrm{e}{-6}$ \\
\end{longtable}
\noindent\footnotesize SD is the sample standard deviation across five training seeds. NF$_{k/5}$ denotes a group with $k$ non-finite seeds. Compact notation $a\mathrm{e}{b}$ means $a\times10^b$.
\endgroup

\begingroup\small
\setlength{\tabcolsep}{2.4pt}
\renewcommand{\arraystretch}{1.15}
\setlength{\LTleft}{\fill}\setlength{\LTright}{\fill}
\setlength{\LTcapwidth}{\linewidth}
\begin{longtable}{@{}lrrrrrrrr@{}}
\caption{\textbf{Nonlinear prediction: full seed statistics.} GMean100 for the nonlinear block of Table~\ref{tab:main-comparison}.}\label{tab:main-nonlinear-seeds}\\
\toprule
Method & \multicolumn{3}{c}{Summary} & \multicolumn{5}{c}{Training run} \\
\cmidrule(lr){2-4}\cmidrule(l){5-9}
 & Median & Mean & SD & A & B & C & D & E \\
\midrule
\endfirsthead
\multicolumn{9}{l}{\tablename\ \thetable\ (continued)}\\
\toprule
Method & \multicolumn{3}{c}{Summary} & \multicolumn{5}{c}{Training run} \\
\cmidrule(lr){2-4}\cmidrule(l){5-9}
 & Median & Mean & SD & A & B & C & D & E \\
\midrule
\endhead
\midrule\multicolumn{9}{r}{\footnotesize Continued on the next page}\\
\endfoot
\bottomrule
\endlastfoot
\addlinespace[5pt]\multicolumn{9}{l}{\textit{KdV (1D)}}\\*
FNO & $0.194$ & $0.188$ & $0.012$ & $0.168$ & $0.187$ & $0.194$ & $0.195$ & $0.196$ \\*
CNO & $0.473$ & $0.469$ & $0.048$ & $0.482$ & $0.402$ & $0.473$ & $0.534$ & $0.455$ \\*
KNO & $0.0908$ & $0.0873$ & $0.0088$ & $0.0729$ & $0.0919$ & $0.0955$ & $0.0908$ & $0.0851$ \\*
IS-FNO(1) & $0.0466$ & $0.0468$ & $0.0039$ & $0.0466$ & $0.0462$ & $0.0521$ & $0.0412$ & $0.0480$ \\*
\textbf{SGNO} & $0.0122$ & $0.0130$ & $0.0028$ & $0.0122$ & $0.0141$ & $0.0172$ & $0.0103$ & $0.0110$ \\
\addlinespace[5pt]\multicolumn{9}{l}{\textit{KS (1D)}}\\*
FNO & $0.516$ & $0.517$ & $0.012$ & $0.520$ & $0.501$ & $0.516$ & $0.515$ & $0.534$ \\*
CNO & $0.360$ & $0.583$ & $0.51$ & $0.371$ & $1.49$ & $0.350$ & $0.342$ & $0.360$ \\*
KNO & $0.307$ & $0.305$ & $0.022$ & $0.277$ & $0.307$ & $0.336$ & $0.292$ & $0.312$ \\*
IS-FNO(1) & $0.0800$ & $0.0781$ & $0.0057$ & $0.0800$ & $0.0699$ & $0.0837$ & $0.0748$ & $0.0820$ \\*
\textbf{SGNO} & $0.0512$ & $0.0565$ & $0.012$ & $0.0771$ & $0.0582$ & $0.0457$ & $0.0504$ & $0.0512$ \\
\addlinespace[5pt]\multicolumn{9}{l}{\textit{Burgers (1D)}}\\*
FNO & $0.0185$ & $0.0186$ & $0.0022$ & $0.0185$ & $0.0166$ & $0.0193$ & $0.0219$ & $0.0167$ \\*
CNO & $0.293$ & $0.289$ & $0.033$ & $0.313$ & $0.293$ & $0.326$ & $0.248$ & $0.263$ \\*
KNO & $0.0196$ & $0.0623$ & $0.097$ & $0.235$ & $0.0224$ & $0.0196$ & $0.0177$ & $0.0167$ \\*
IS-FNO(1) & $0.00584$ & $0.00585$ & $4.1\mathrm{e}{-4}$ & $0.00521$ & $0.00584$ & $0.00603$ & $0.00635$ & $0.00582$ \\*
\textbf{SGNO} & $0.00334$ & $0.00345$ & $4.2\mathrm{e}{-4}$ & $0.00322$ & $0.00302$ & $0.00334$ & $0.00353$ & $0.00412$ \\
\addlinespace[5pt]\multicolumn{9}{l}{\textit{Kolmogorov (2D)}}\\*
FNO & $0.666$ & $0.653$ & $0.029$ & $0.670$ & $0.600$ & $0.659$ & $0.668$ & $0.666$ \\*
CNO & $0.494$ & $0.488$ & $0.016$ & $0.494$ & $0.494$ & $0.465$ & $0.508$ & $0.480$ \\*
KNO & $0.976$ & $0.980$ & $0.034$ & $0.950$ & $1.03$ & $0.976$ & $0.994$ & $0.949$ \\*
IS-FNO(1) & $0.552$ & $0.546$ & $0.015$ & $0.536$ & $0.561$ & $0.526$ & $0.558$ & $0.552$ \\*
\textbf{SGNO} & $0.430$ & $0.441$ & $0.035$ & $0.468$ & $0.404$ & $0.418$ & $0.430$ & $0.487$ \\
\end{longtable}
\noindent\footnotesize SD is the sample standard deviation across five training seeds. NF$_{k/5}$ denotes a group with $k$ non-finite seeds. Compact notation $a\mathrm{e}{b}$ means $a\times10^b$.
\endgroup

\FloatBarrier

\subsection{Linear update recovery}
\label{app:linear_recovery}
Table~\ref{tab:linear-fitted-only} compares the fitted map alone with the
jointly trained predictor. Both reach small errors on these linear tasks;
the comparison identifies the contribution of fitting the state update
before neural training.

\begin{table}[!htbp]
\centering\small
\caption{\textbf{Fitted-only and jointly trained linear predictions.} GMean100; both columns are scaled by $10^{-6}$. Fitted-only is deterministic. SGNO reports the five-seed median after joint training.}
\label{tab:linear-fitted-only}
\setlength{\tabcolsep}{8pt}\renewcommand{\arraystretch}{1.15}
\begin{tabular*}{\linewidth}{@{\extracolsep{\fill}}lrr@{}}
\toprule
Linear task & Fitted-only & SGNO \\
\midrule
Advection (1D) & $6.20$ & $2.54$ \\
Diffusion (1D) & $7.21$ & $8.12$ \\
Dispersion (1D) & $7.24$ & $5.01$ \\
Mixed dispersion (2D) & $7.70$ & $6.06$ \\
Axis diffusion (2D) & $8.53$ & $7.74$ \\
Rotated diffusion (2D) & $8.35$ & $7.57$ \\
Diagonal diffusion (3D) & $12.88$ & $5.80$ \\
Unbalanced advection (3D) & $16.00$ & $5.77$ \\
Advection--diffusion (3D) & $11.79$ & $7.11$ \\
\bottomrule
\end{tabular*}
\end{table}

\FloatBarrier

\subsection{Additional training objectives and capacity}
\label{app:additional_protocols}\label{app:modal_fits}
Table~\ref{tab:training-protocol-controls} compares complete reported
task groups under additional training choices. FNO(20) unrolls 20
physical-state predictions; IS-FNO(20) predicts 20 states, both using
relative-$\ell_2$ supervision within the 50-transition training records.
Both use 10,000 updates and effective batch size 20. The added FNO runs
use Adam at $10^{-3}$; the Kolmogorov IS-FNO run uses $10^{-3}$,
weight decay $10^{-6}$, $\epsilon=10^{-6}$, and gradient clipping at 10.
PDE-Refiner-L uses a normalized width-64 U-Net (53.76M array elements),
three refinement stages, denoising supervision and an exponential moving
average (EMA), trained for
50,000 updates with peak rate $10^{-4}$, 2,000 warm-up updates,
cosine decay, and effective batch size 20. Its capacity and budget are
separate from the main comparison. Extra temporal targets and refinement
calls make equal optimizer-update counts different compute budgets.

\begin{table}[!htbp]
\centering
\caption{\textbf{Additional objectives and model capacity.} GMean100,
five-seed mean $\pm$ sample SD. SGNO, FNO(20), and IS-FNO(20) use 10k
updates; Refiner-L uses 53.76M parameter-array elements and 50k updates. ---: not
evaluated; NF$_{2/5}$: two non-finite seeds in steps 1--100.}
\label{tab:training-protocol-controls}
\begingroup\small
\setlength{\tabcolsep}{3pt}\renewcommand{\arraystretch}{1.15}
\begin{tabular*}{\linewidth}{@{\extracolsep{\fill}}lrrrr@{}}
\toprule
Task & SGNO & FNO(20) & IS-FNO(20) & PDE-Refiner-L \\
\midrule
KdV & \ms{0.0130}{0.0028} & \ms{0.163}{0.016} & \ms{0.0326}{0.0015} & --- \\
KS & \ms{0.0565}{0.012} & \ms{0.843}{0.028} & \ms{0.127}{0.018} & \ms{0.00560}{0.00079} \\
Burgers & \ms{0.00345}{0.00042} & \ms{0.0244}{0.0043} & \ms{0.00382}{0.00028} & \ms{0.00332}{0.0015} \\
Kolmogorov & \ms{0.441}{0.035} & \ms{0.744}{0.015} & $\mathrm{NF}_{2/5}$ & --- \\
\bottomrule
\end{tabular*}\endgroup
\end{table}

\FloatBarrier

\section{Component-ablation results}
\label{app:ablation_results}
Tables~\ref{tab:ablation-all-seeds} and~\ref{tab:ablation-path-seeds}
expand Table~\ref{tab:main-ablation}. They separate the matched
direct-complex initialization controls from the original SGNO
parameterization and its layer-path interventions.
Appendix~\ref{app:component_protocol} defines both protocols and their
evaluated task groups.

\subsection{Initialization and joint adaptation}
\begingroup\small
\setlength{\tabcolsep}{2.4pt}
\renewcommand{\arraystretch}{1.15}
\setlength{\LTleft}{\fill}\setlength{\LTright}{\fill}
\setlength{\LTcapwidth}{\linewidth}
\begin{longtable}{@{}lrrrrrrrr@{}}
\caption{\textbf{Initialization and adaptation controls.} Per-seed GMean100 and summary statistics for the matched direct-complex controls in Table~\ref{tab:main-ablation}.}\label{tab:ablation-all-seeds}\\
\toprule
Method & \multicolumn{3}{c}{Summary} & \multicolumn{5}{c}{Training run} \\
\cmidrule(lr){2-4}\cmidrule(l){5-9}
 & Median & Mean & SD & A & B & C & D & E \\
\midrule
\endfirsthead
\multicolumn{9}{l}{\tablename\ \thetable\ (continued)}\\
\toprule
Method & \multicolumn{3}{c}{Summary} & \multicolumn{5}{c}{Training run} \\
\cmidrule(lr){2-4}\cmidrule(l){5-9}
 & Median & Mean & SD & A & B & C & D & E \\
\midrule
\endhead
\midrule\multicolumn{9}{r}{\footnotesize Continued on the next page}\\
\endfoot
\bottomrule
\endlastfoot
\addlinespace[5pt]\multicolumn{9}{l}{\textit{KdV (1D)}}\\*
Fixed identity & $0.0165$ & $0.0168$ & $0.0023$ & $0.0138$ & $0.0165$ & $0.0153$ & $0.0187$ & $0.0194$ \\*
Trainable identity & $0.0143$ & $0.0139$ & $0.0016$ & $0.0117$ & $0.0143$ & $0.0128$ & $0.0155$ & $0.0154$ \\*
Fixed fitted map & $0.0129$ & $0.0128$ & $0.0014$ & $0.0129$ & $0.0141$ & $0.0140$ & $0.0109$ & $0.0120$ \\*
Trainable fitted & $0.0121$ & $0.0131$ & $0.0029$ & $0.0121$ & $0.0133$ & $0.0179$ & $0.0102$ & $0.0118$ \\
\addlinespace[5pt]\multicolumn{9}{l}{\textit{KS (1D)}}\\*
Fixed identity & $0.0541$ & $0.0512$ & $0.011$ & $0.0541$ & $0.0591$ & $0.0588$ & $0.0511$ & $0.0330$ \\*
Trainable identity & $0.0498$ & $0.0522$ & $0.010$ & $0.0624$ & $0.0632$ & $0.0406$ & $0.0450$ & $0.0498$ \\*
Fixed fitted map & $0.0529$ & $0.0562$ & $0.0090$ & $0.0632$ & $0.0679$ & $0.0466$ & $0.0529$ & $0.0505$ \\*
Trainable fitted & $0.0477$ & $0.0527$ & $0.012$ & $0.0642$ & $0.0658$ & $0.0400$ & $0.0461$ & $0.0477$ \\
\addlinespace[5pt]\multicolumn{9}{l}{\textit{Burgers (1D)}}\\*
Fixed identity & $0.00363$ & $0.00364$ & $3.6\mathrm{e}{-4}$ & $0.00338$ & $0.00363$ & $0.00322$ & $0.00389$ & $0.00409$ \\*
Trainable identity & $0.00356$ & $0.00355$ & $5.8\mathrm{e}{-4}$ & $0.00380$ & $0.00356$ & $0.00263$ & $0.00353$ & $0.00421$ \\*
Fixed fitted map & $0.00368$ & $0.00373$ & $5.8\mathrm{e}{-4}$ & $0.00283$ & $0.00368$ & $0.00363$ & $0.00422$ & $0.00427$ \\*
Trainable fitted & $0.00338$ & $0.00360$ & $5.5\mathrm{e}{-4}$ & $0.00327$ & $0.00323$ & $0.00338$ & $0.00359$ & $0.00455$ \\
\addlinespace[5pt]\multicolumn{9}{l}{\textit{Kolmogorov (2D)}}\\*
Fixed identity & $0.493$ & $0.484$ & $0.035$ & $0.493$ & $0.454$ & $0.528$ & $0.500$ & $0.442$ \\*
Trainable identity & $0.445$ & $0.451$ & $0.030$ & $0.416$ & $0.445$ & $0.497$ & $0.461$ & $0.435$ \\*
Fixed fitted map & $0.480$ & $0.533$ & $0.13$ & $0.755$ & $0.424$ & $0.480$ & $0.472$ & $0.534$ \\*
Trainable fitted & $0.427$ & $0.446$ & $0.042$ & $0.491$ & $0.421$ & $0.400$ & $0.427$ & $0.492$ \\
\end{longtable}
\noindent\footnotesize The four settings share the correction initialization, training data order, loss, output processing, and training budget. SD denotes sample standard deviation.
\endgroup

\subsection{Correction-layer paths}
\begingroup\small
\setlength{\tabcolsep}{2.4pt}
\renewcommand{\arraystretch}{1.15}
\setlength{\LTleft}{\fill}\setlength{\LTright}{\fill}
\setlength{\LTcapwidth}{\linewidth}
\begin{longtable}{@{}lrrrrrrrr@{}}
\caption{\textbf{Correction-layer path controls.} Per-seed GMean100 and summary statistics. These runs use the original outer-map parameterization, separately from Table~\ref{tab:ablation-all-seeds}.}\label{tab:ablation-path-seeds}\\
\toprule
Method & \multicolumn{3}{c}{Summary} & \multicolumn{5}{c}{Training run} \\
\cmidrule(lr){2-4}\cmidrule(l){5-9}
 & Median & Mean & SD & A & B & C & D & E \\
\midrule
\endfirsthead
\multicolumn{9}{l}{\tablename\ \thetable\ (continued)}\\
\toprule
Method & \multicolumn{3}{c}{Summary} & \multicolumn{5}{c}{Training run} \\
\cmidrule(lr){2-4}\cmidrule(l){5-9}
 & Median & Mean & SD & A & B & C & D & E \\
\midrule
\endhead
\midrule\multicolumn{9}{r}{\footnotesize Continued on the next page}\\
\endfoot
\bottomrule
\endlastfoot
\addlinespace[5pt]\multicolumn{9}{l}{\textit{KdV (1D)}}\\*
\textbf{SGNO} & $0.0122$ & $0.0130$ & $0.0028$ & $0.0122$ & $0.0141$ & $0.0172$ & $0.0103$ & $0.0110$ \\*
No spectral source & $173$ & $9.89\mathrm{e}{13}$ & $2.2\mathrm{e}{14}$ & $4.95\mathrm{e}{14}$ & $1.61$ & $173$ & $1.34$ & $2.97\mathrm{e}{3}$ \\*
No affine path & $0.0105$ & $0.0110$ & $0.0027$ & $0.00912$ & $0.0105$ & $0.0109$ & $0.0157$ & $0.00889$ \\
\addlinespace[5pt]\multicolumn{9}{l}{\textit{KS (1D)}}\\*
\textbf{SGNO} & $0.0512$ & $0.0565$ & $0.012$ & $0.0771$ & $0.0582$ & $0.0457$ & $0.0504$ & $0.0512$ \\*
No spectral source & $0.241$ & $0.464$ & $0.52$ & $0.241$ & $0.241$ & $0.236$ & $0.219$ & $1.39$ \\*
No affine path & $0.0640$ & $0.0773$ & $0.028$ & $0.109$ & $0.0573$ & $0.106$ & $0.0640$ & $0.0499$ \\
\addlinespace[5pt]\multicolumn{9}{l}{\textit{Burgers (1D)}}\\*
\textbf{SGNO} & $0.00334$ & $0.00345$ & $4.2\mathrm{e}{-4}$ & $0.00322$ & $0.00302$ & $0.00334$ & $0.00353$ & $0.00412$ \\
\addlinespace[5pt]\multicolumn{9}{l}{\textit{Kolmogorov (2D)}}\\*
\textbf{SGNO} & $0.430$ & $0.441$ & $0.035$ & $0.468$ & $0.404$ & $0.418$ & $0.430$ & $0.487$ \\
\end{longtable}
\noindent\footnotesize Path removals were evaluated on KdV and KS. Only the complete SGNO model was evaluated in this group on Burgers and Kolmogorov. The removed affine path is inside each correction layer.
\endgroup

\subsection{Output processing}
\label{app:output_processing}
Table~\ref{tab:output-processing} reports retrained SGNO controls with
the output filter, high-frequency zero-input subtraction, or both removed.
The fitted/trainable outer map, mean handling, and training protocol are
retained. Removing both changes task medians by less than 3.8\% relative
to the complete-processing controls. These two operations have a small
effect on prediction error under this protocol.

\begingroup\small
\setlength{\tabcolsep}{2.3pt}\renewcommand{\arraystretch}{1.1}
\setlength{\LTleft}{\fill}\setlength{\LTright}{\fill}\setlength{\LTcapwidth}{\linewidth}
\begin{longtable}{@{}lrrrrrrrr@{}}
\caption{\textbf{SGNO output-processing controls.} GMean100 after retraining with the output filter, high-frequency zero-input subtraction, or both removed. All values are unscaled; SD is the sample standard deviation.}\label{tab:output-processing}\\
\toprule Setting & Median & Mean & SD & \multicolumn{5}{c}{Training run} \\
\cmidrule(l){5-9} & & & & A & B & C & D & E \\ \midrule
\endfirsthead
\multicolumn{9}{l}{\tablename\ \thetable\ (continued)}\\
\toprule Setting & Median & Mean & SD & A & B & C & D & E \\ \midrule
\endhead
\midrule\multicolumn{9}{r}{Continued on the next page}\\\endfoot
\bottomrule\endlastfoot
\addlinespace[4pt]\multicolumn{9}{l}{\textit{KdV}}\\*
Complete & $0.0122$ & $0.0129$ & $0.00264$ & $0.0122$ & $0.0131$ & $0.0172$ & $0.0103$ & $0.0115$ \\
No filter & $0.0122$ & $0.0129$ & $0.00263$ & $0.0122$ & $0.0131$ & $0.0172$ & $0.0103$ & $0.0116$ \\
No subtraction & $0.0124$ & $0.0132$ & $0.00306$ & $0.0124$ & $0.0134$ & $0.0184$ & $0.0106$ & $0.0113$ \\
Neither & $0.0124$ & $0.0132$ & $0.00312$ & $0.0124$ & $0.0134$ & $0.0185$ & $0.0106$ & $0.0113$ \\
\addlinespace[4pt]\multicolumn{9}{l}{\textit{KS}}\\*
Complete & $0.0512$ & $0.0566$ & $0.0124$ & $0.0772$ & $0.0586$ & $0.0457$ & $0.05$ & $0.0512$ \\
No filter & $0.0512$ & $0.0565$ & $0.0123$ & $0.0769$ & $0.0586$ & $0.0456$ & $0.05$ & $0.0512$ \\
No subtraction & $0.0514$ & $0.0563$ & $0.012$ & $0.076$ & $0.0588$ & $0.0453$ & $0.0501$ & $0.0514$ \\
Neither & $0.0516$ & $0.0563$ & $0.0119$ & $0.0758$ & $0.0586$ & $0.0453$ & $0.0501$ & $0.0516$ \\
\addlinespace[4pt]\multicolumn{9}{l}{\textit{Burgers}}\\*
Complete & $0.00334$ & $0.00357$ & $0.000563$ & $0.00321$ & $0.00313$ & $0.00334$ & $0.00369$ & $0.00451$ \\
No filter & $0.00333$ & $0.00358$ & $0.000561$ & $0.00323$ & $0.00313$ & $0.00333$ & $0.00369$ & $0.0045$ \\
No subtraction & $0.00346$ & $0.0036$ & $0.000468$ & $0.00326$ & $0.00315$ & $0.00346$ & $0.00379$ & $0.00431$ \\
Neither & $0.00346$ & $0.00359$ & $0.000472$ & $0.00325$ & $0.00315$ & $0.00346$ & $0.0038$ & $0.00431$ \\
\addlinespace[4pt]\multicolumn{9}{l}{\textit{Kolmogorov}}\\*
Complete & $0.43$ & $0.441$ & $0.0343$ & $0.467$ & $0.404$ & $0.418$ & $0.43$ & $0.486$ \\
No filter & $0.429$ & $0.441$ & $0.0334$ & $0.467$ & $0.405$ & $0.418$ & $0.429$ & $0.485$ \\
No subtraction & $0.43$ & $0.443$ & $0.0386$ & $0.465$ & $0.402$ & $0.418$ & $0.43$ & $0.498$ \\
Neither & $0.43$ & $0.448$ & $0.0466$ & $0.466$ & $0.403$ & $0.419$ & $0.43$ & $0.52$ \\
\end{longtable}
\endgroup

\FloatBarrier

\section{Long-horizon prediction results}
\label{app:current_temporal}
Table~\ref{tab:temporal-compact} reports GMean101--200, step-200 nRMSE,
and absolute nRMSE slopes over steps 101--200 for all four nonlinear
tasks. Each training run first averages nRMSE over the test initial
states. The table reports all five runs, together with their median,
mean, and sample SD.

For one training run, let $\bar e_t$ denote the test-mean nRMSE at
prediction step $t$. Define the mean step index
$\bar t=150.5$ and the mean error within the late window as
$\bar e=100^{-1}\sum_{t=101}^{200}\bar e_t$.
The ordinary least-squares slope is
\begin{equation}
\beta_e =
\frac{
\sum_{t=101}^{200}(t-\bar t)(\bar e_t-\bar e)
}{
\sum_{t=101}^{200}(t-\bar t)^2
}.
\label{eq:late_absolute_slope}
\end{equation}
This slope measures the absolute change in nRMSE per prediction step
within the evaluated window. Table~\ref{tab:temporal-compact} reports
slope values in units of $10^{-3}$ per step. The fit is performed
separately for each training run before computing summary statistics.

On KdV, the lower endpoint median (0.0625 versus 0.136) coexists with
a higher mean (0.344 versus 0.137), reflecting the larger between-run
variation visible in the table.

\begingroup
\small
\setlength{\tabcolsep}{2.4pt}
\renewcommand{\arraystretch}{1.15}
\setlength{\LTleft}{\fill}
\setlength{\LTright}{\fill}
\setlength{\LTcapwidth}{\linewidth}

\begin{longtable}{@{}lrrrrrrrr@{}}
\caption{\textbf{Long-horizon prediction errors and absolute error slopes.}
GMean101--200, endpoint nRMSE, and the ordinary least-squares slope
$\beta_e$ of nRMSE over steps 101--200.
Each run first averages nRMSE over all test initial states.
Summary statistics use all five training runs.
Slope entries are in units of $10^{-3}$ per prediction step.}
\label{tab:temporal-compact}\\

\toprule
Method & \multicolumn{3}{c}{Summary}
       & \multicolumn{5}{c}{Training run} \\
\cmidrule(lr){2-4}\cmidrule(l){5-9}
 & Median & Mean & SD & A & B & C & D & E \\
\midrule
\endfirsthead

\multicolumn{9}{l}{\tablename\ \thetable\ (continued)}\\
\toprule
Method & \multicolumn{3}{c}{Summary}
       & \multicolumn{5}{c}{Training run} \\
\cmidrule(lr){2-4}\cmidrule(l){5-9}
 & Median & Mean & SD & A & B & C & D & E \\
\midrule
\endhead

\midrule
\multicolumn{9}{r}{\footnotesize Continued on the next page}\\
\endfoot

\bottomrule
\endlastfoot

\addlinespace[5pt]
\multicolumn{9}{l}{\textit{KdV (1D): GMean101--200}}\\*
IS-FNO(1)
 & $0.108$ & $0.108$ & $0.0098$
 & $0.113$ & $0.104$ & $0.108$ & $0.0954$ & $0.121$ \\*
\textbf{SGNO}
 & $0.0418$ & $0.0913$ & $0.11$
 & $0.0397$ & $0.0519$ & $0.288$ & $0.0351$ & $0.0418$ \\

\addlinespace[5pt]
\multicolumn{9}{l}{\textit{KdV (1D): nRMSE at step 200}}\\*
IS-FNO(1)
 & $0.136$ & $0.137$ & $0.019$
 & $0.153$ & $0.136$ & $0.122$ & $0.115$ & $0.160$ \\*
\textbf{SGNO}
 & $0.0625$ & $0.344$ & $0.64$
 & $0.0528$ & $0.0699$ & $1.48$ & $0.0509$ & $0.0625$ \\

\addlinespace[5pt]
\multicolumn{9}{l}{
\textit{KdV (1D): Absolute-error slope $\beta_e$ ($\times10^{-3}$)}
}\\*
IS-FNO(1)
 & $0.575$ & $0.530$ & $0.19$
 & $0.735$ & $0.575$ & $0.289$ & $0.392$ & $0.658$ \\*
\textbf{SGNO}
 & $0.318$ & $2.71$ & $5.4$
 & $0.242$ & $0.318$ & $12.3$ & $0.283$ & $0.378$ \\

\addlinespace[5pt]
\multicolumn{9}{l}{\textit{KS (1D): GMean101--200}}\\*
IS-FNO(1)
 & $0.715$ & $0.704$ & $0.045$
 & $0.749$ & $0.633$ & $0.715$ & $0.731$ & $0.693$ \\*
\textbf{SGNO}
 & $0.491$ & $0.510$ & $0.070$
 & $0.619$ & $0.531$ & $0.436$ & $0.491$ & $0.472$ \\

\addlinespace[5pt]
\multicolumn{9}{l}{\textit{KS (1D): nRMSE at step 200}}\\*
IS-FNO(1)
 & $1.07$ & $1.07$ & $0.056$
 & $1.07$ & $1.01$ & $1.11$ & $1.14$ & $1.02$ \\*
\textbf{SGNO}
 & $0.830$ & $0.841$ & $0.062$
 & $0.937$ & $0.860$ & $0.774$ & $0.830$ & $0.804$ \\

\addlinespace[5pt]
\multicolumn{9}{l}{
\textit{KS (1D): Absolute-error slope $\beta_e$ ($\times10^{-3}$)}
}\\*
IS-FNO(1)
 & $7.06$ & $7.15$ & $0.60$
 & $6.55$ & $7.06$ & $7.67$ & $7.84$ & $6.61$ \\*
\textbf{SGNO}
 & $6.01$ & $6.03$ & $0.20$
 & $5.99$ & $6.30$ & $5.74$ & $6.10$ & $6.01$ \\

\addlinespace[5pt]
\multicolumn{9}{l}{\textit{Burgers (1D): GMean101--200}}\\*
IS-FNO(1)
 & $0.0155$ & $0.0154$ & $8.9\mathrm{e}{-4}$
 & $0.0139$ & $0.0155$ & $0.0155$ & $0.0161$ & $0.0159$ \\*
\textbf{SGNO}
 & $0.00797$ & $0.00864$ & $0.0014$
 & $0.00858$ & $0.00797$ & $0.00764$ & $0.00794$ & $0.0111$ \\

\addlinespace[5pt]
\multicolumn{9}{l}{\textit{Burgers (1D): nRMSE at step 200}}\\*
IS-FNO(1)
 & $0.0215$ & $0.0216$ & $0.0014$
 & $0.0199$ & $0.0216$ & $0.0209$ & $0.0215$ & $0.0238$ \\*
\textbf{SGNO}
 & $0.0118$ & $0.0122$ & $0.0023$
 & $0.0118$ & $0.0118$ & $0.0105$ & $0.0108$ & $0.0161$ \\

\addlinespace[5pt]
\multicolumn{9}{l}{
\textit{Burgers (1D): Absolute-error slope $\beta_e$ ($\times10^{-3}$)}
}\\*
IS-FNO(1)
 & $0.104$ & $0.108$ & $0.014$
 & $0.104$ & $0.107$ & $0.0988$ & $0.0987$ & $0.133$ \\*
\textbf{SGNO}
 & $0.0615$ & $0.0631$ & $0.015$
 & $0.0615$ & $0.0660$ & $0.0514$ & $0.0499$ & $0.0869$ \\

\addlinespace[5pt]
\multicolumn{9}{l}{\textit{Kolmogorov (2D): GMean101--200}}\\*
IS-FNO(1)
 & $\mathrm{NF}_{1/5}$ & $\mathrm{NF}_{1/5}$ & $\mathrm{NF}_{1/5}$
 & $\mathrm{NF}$ & $1.21$ & $1.17$ & $1.21$ & $1.18$ \\*
\textbf{SGNO}
 & $1.12$ & $1.12$ & $0.021$
 & $1.13$ & $1.10$ & $1.10$ & $1.12$ & $1.15$ \\

\addlinespace[5pt]
\multicolumn{9}{l}{\textit{Kolmogorov (2D): nRMSE at step 200}}\\*
IS-FNO(1)
 & $\mathrm{NF}_{1/5}$ & $\mathrm{NF}_{1/5}$ & $\mathrm{NF}_{1/5}$
 & $\mathrm{NF}$ & $1.23$ & $1.18$ & $1.26$ & $1.19$ \\*
\textbf{SGNO}
 & $1.17$ & $1.17$ & $0.019$
 & $1.19$ & $1.14$ & $1.18$ & $1.17$ & $1.16$ \\

\addlinespace[5pt]
\multicolumn{9}{l}{
\textit{Kolmogorov (2D): Absolute-error slope $\beta_e$ ($\times10^{-3}$)}
}\\*
IS-FNO(1)
 & $\mathrm{NF}_{1/5}$ & $\mathrm{NF}_{1/5}$ & $\mathrm{NF}_{1/5}$
 & $\mathrm{NF}$ & $0.805$ & $0.796$ & $1.46$ & $0.567$ \\*
\textbf{SGNO}
 & $1.31$ & $1.33$ & $0.34$
 & $1.25$ & $1.37$ & $1.82$ & $1.31$ & $0.882$ \\

\end{longtable}

\noindent
\footnotesize
SD is the sample standard deviation across five training runs.
Columns A--E identify the paired training seeds across methods.
$\mathrm{NF}_{k/5}$ denotes a group containing $k$ non-finite runs;
group summaries are not computed after excluding these runs.
Compact notation $a\mathrm{e}{b}$ means $a\times10^b$.
\endgroup
\FloatBarrier

\section{Spectral and field diagnostics}
\label{app:spectral_diagnostics}

\subsection{Amplitude and phase accounting}
Write predicted and reference Fourier coefficients as
$\rho e^{i\phi}$ and $r e^{i\psi}$, with magnitudes $\rho,r\geq0$
and phases $\phi,\psi$. Then
\begin{equation}
|\rho e^{i\phi}-r e^{i\psi}|^2
=(\rho-r)^2+2\rho r[1-\cos(\phi-\psi)].
\label{eq:phase-magnitude-split}
\end{equation}
The first term measures amplitude error. The second measures phase
differences weighted by the predicted and reference magnitudes.
We sum both terms using real-FFT Parseval weights and normalize by
reference energy, obtaining two nonnegative contributions whose sum
equals the normalized squared field error.

Figure~\ref{fig:mechanism-summary} compares IS-FNO(1) and SGNO on KS
at step 200. Each plotted point is the average normalized squared
component over 30 test initial states for one training run.
No square root is taken. Diamonds show the mean across five runs,
and error bars show the sample standard deviation, computed with
denominator $5-1$. Gray lines connect runs with the same training
seed across the two models; all five runs are included.

\subsection{Field norms and correlation}
\label{app:field_diagnostics}
For a nonzero reference field $\mathbf u$ and nonzero predicted field
$\widetilde{\mathbf u}$ at step 200, define the predicted-to-reference
squared-norm ratio
\begin{equation}
\nu =
\frac{\|\widetilde{\mathbf u}\|_2^2}{\|\mathbf u\|_2^2}
\label{eq:field_norm_ratio}
\end{equation}
and the field correlation
\begin{equation}
c_{\mathrm{field}} =
\frac{
\langle\widetilde{\mathbf u},\mathbf u\rangle
}{
\|\widetilde{\mathbf u}\|_2\|\mathbf u\|_2
}.
\label{eq:field_correlation}
\end{equation}
Here $\langle\cdot,\cdot\rangle$ is the Euclidean inner product.
A squared-norm ratio of one indicates matching field magnitude;
a correlation of one indicates aligned field vectors.
These quantities are related to the relative squared prediction error by
\begin{equation}
\frac{\|\widetilde{\mathbf u}-\mathbf u\|_2^2}{\|\mathbf u\|_2^2}
=
1+\nu-2\sqrt{\nu}\,c_{\mathrm{field}}.
\label{eq:field_norm_correlation}
\end{equation}
This identity holds for each predicted--reference field pair before
aggregation.

For KS, we compute both diagnostics for each of the 30 test initial
states and average within each training run.
Table~\ref{tab:ks-field-endpoint} reports all five run-level values,
their median, mean, and sample SD. Section~\ref{subsec:propagation_mechanism}
reports the medians: SGNO's squared-norm ratio is closer to one
(1.05 versus 1.14), and its field correlation is higher
(0.601 versus 0.402).

\begin{table}[!htbp]
\centering
\caption{\textbf{KS field diagnostics at step 200.}
Each run averages over all 30 test initial states.
A squared-norm ratio closer to one and a higher field correlation
indicate better agreement with the reference.
SD is the sample standard deviation across five training runs;
columns A--E identify paired runs across methods.}
\label{tab:ks-field-endpoint}
\begingroup
\small
\setlength{\tabcolsep}{2.4pt}
\renewcommand{\arraystretch}{1.13}
\begin{tabular*}{\linewidth}{@{\extracolsep{\fill}}lrrrrrrrr@{}}
\toprule
Method & \multicolumn{3}{c}{Summary}
       & \multicolumn{5}{c}{Training run} \\
\cmidrule(lr){2-4}\cmidrule(l){5-9}
 & Median & Mean & SD & A & B & C & D & E \\
\midrule
\multicolumn{9}{l}{\textit{Squared-norm ratio $\nu$}} \\
IS-FNO(1)
 & $1.14$ & $1.15$ & $0.0907$
 & $1.14$ & $1.13$ & $1.16$ & $1.29$ & $1.04$ \\
\textbf{SGNO}
 & $1.05$ & $1.04$ & $0.0470$
 & $1.09$ & $1.07$ & $1.05$ & $0.969$ & $1.02$ \\
\addlinespace[4pt]
\multicolumn{9}{l}{\textit{Field correlation $c_{\mathrm{field}}$}} \\
IS-FNO(1)
 & $0.402$ & $0.424$ & $0.0538$
 & $0.402$ & $0.482$ & $0.361$ & $0.397$ & $0.479$ \\
\textbf{SGNO}
 & $0.601$ & $0.601$ & $0.0525$
 & $0.529$ & $0.578$ & $0.667$ & $0.601$ & $0.632$ \\
\bottomrule
\end{tabular*}
\endgroup
\end{table}

\FloatBarrier
\section{Additional field visualizations}
\label{app:volume_analysis}
\begin{figure}[!htbp]
\centering
\includegraphics[width=\linewidth]{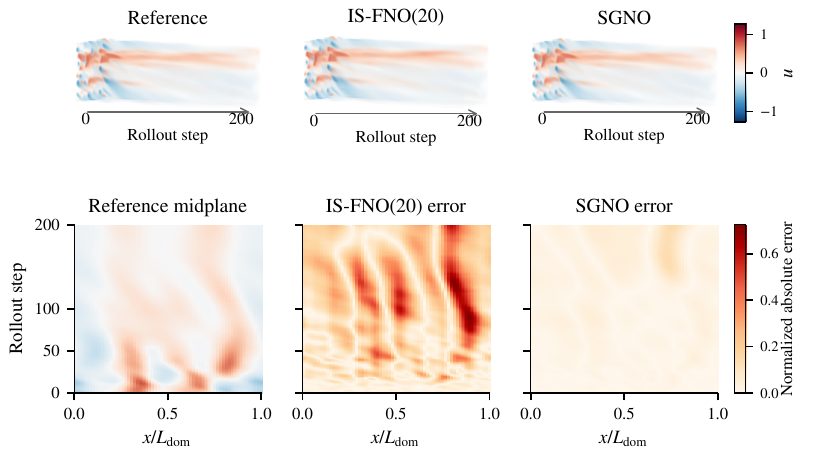}
\caption{\textbf{Space--time structure in the 2D example.}
Top: Exponax/vAPE4D renderings of the reference, IS-FNO(20), and SGNO.
Bottom: the reference field and absolute errors at $y=L_{\mathrm{dom}}/2$.
Errors are normalized by the full reference spatial RMS at each step
and share a color scale.
This native-$64^2$ example matches the extra 2D KdV-type
case in Figure~\ref{fig:intro-motivation}(c); IS-FNO(20) uses its
20-state objective (Appendix~\ref{app:phase_example}).}
\label{fig:apebench-volume}
\end{figure}

Figure~\ref{fig:apebench-volume} expands the 2D KdV-type example from
Figure~\ref{fig:intro-motivation}(c), using the same saved predictions.
The volumes show evolving spatial structure; the fixed midplane locates
error accumulation over time. Both models use the same camera, transfer
function, and state scale; the slice errors also share a scale.

\end{document}